\documentclass[runningheads]{llncs}

\usepackage{comment}
\usepackage{epsfig}
\usepackage{graphicx}
\usepackage{amsmath}
\usepackage{amssymb} 
\usepackage{color}

\usepackage{booktabs}
\usepackage[inline]{enumitem}

\usepackage{algorithm}
\usepackage{algorithmic}
\usepackage{pifont}
\usepackage{etoolbox}
\usepackage{tikz,subfigure}
\usetikzlibrary{arrows,automata}
\usepackage{xr}
\begin{document}
\pagestyle{headings}
\mainmatter
\def\ECCVSubNumber{2382}  


\title{Fair DARTS: Eliminating Unfair Advantages in Differentiable Architecture Search}

\titlerunning{Fair DARTS}
%
\author{Xiangxiang Chu\inst{1}\orcidID{0000-0003-2548-0605}  \and
Tianbao Zhou\inst{2}\orcidID{0000-0002-2133-059X}\thanks{Equal Contribution.}  \and
Bo Zhang\inst{1}\orcidID{0000-0003-0564-617X}$^\star$  \and
Jixiang Li\inst{1}\orcidID{0000-0001-5949-1498}}
\authorrunning{Chu et al.}
%
\institute{Xiaomi AI Lab \\ \email{\{chuxiangxiang,zhangbo11,lijixiang\}@xiaomi.com} \and Minzu University of China \\\email{tianbaochou@163.com}}
%

\maketitle

\begin{abstract}
Differentiable Architecture Search (DARTS) is now a widely disseminated weight-sharing neural architecture search method. However, it suffers from well-known performance collapse due to an inevitable aggregation of skip connections. In this paper, we first disclose that its root cause lies in an \textbf{unfair advantage} in  \textbf{exclusive competition}. Through experiments, we show that if either of two conditions is broken, the collapse disappears. Thereby, we present a novel approach called Fair DARTS where the exclusive competition is relaxed to  be collaborative. Specifically, we let each operation's architectural weight be independent of others. Yet there is still an important issue of discretization discrepancy. We then propose a \textbf{zero-one} loss to push architectural weights towards zero or one, which approximates an expected multi-hot solution. Our experiments are performed on two mainstream search spaces, and we derive new state-of-the-art results on CIFAR-10 and ImageNet\footnote{Code is available here: \url{https://github.com/xiaomi-automl/FairDARTS}}.




\end{abstract}


\section{Introduction}\label{sec:intro}
In the wake of the DARTS's open-sourcing \cite{liu2018darts}, a diverse number of its variants emerge in the \emph{neural architecture search} community. Some of them extend its use in higher-level architecture search spaces with performance awareness in mind \cite{cai2018proxylessnas,wu2018fbnet}, some learn a stochastic distribution instead of architectural parameters \cite{wu2018fbnet,xie2018snas,zheng2019multinomial,dong2019one,dong2019searching}, and others offer remedies on discovering its lack of robustness \cite{nayman2019xnas,chen2019progressive,liang2019darts,li2019stacnas,zela2020understanding}.

\begin{figure}[ht]
	\centering
	\includegraphics[scale=0.5]{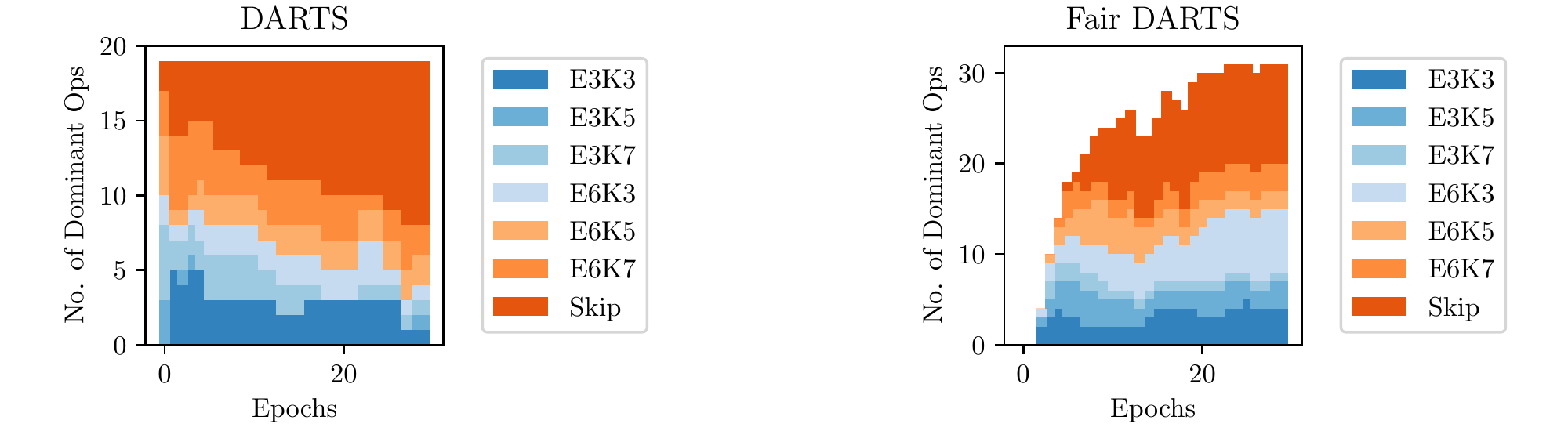}
	\includegraphics[width=0.45\textwidth,scale=0.6]{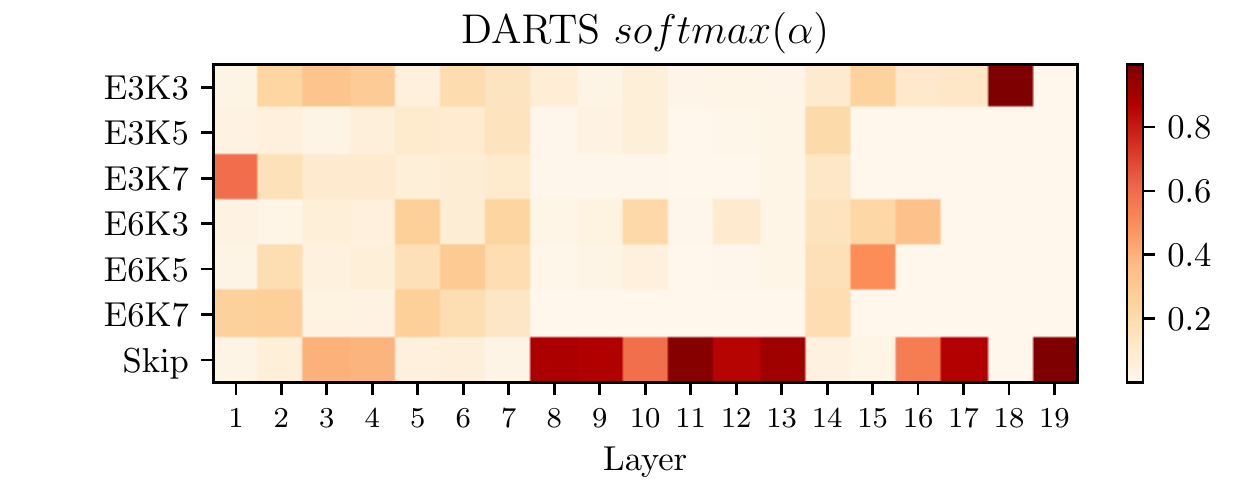}
	\includegraphics[width=0.45\textwidth,scale=0.6]{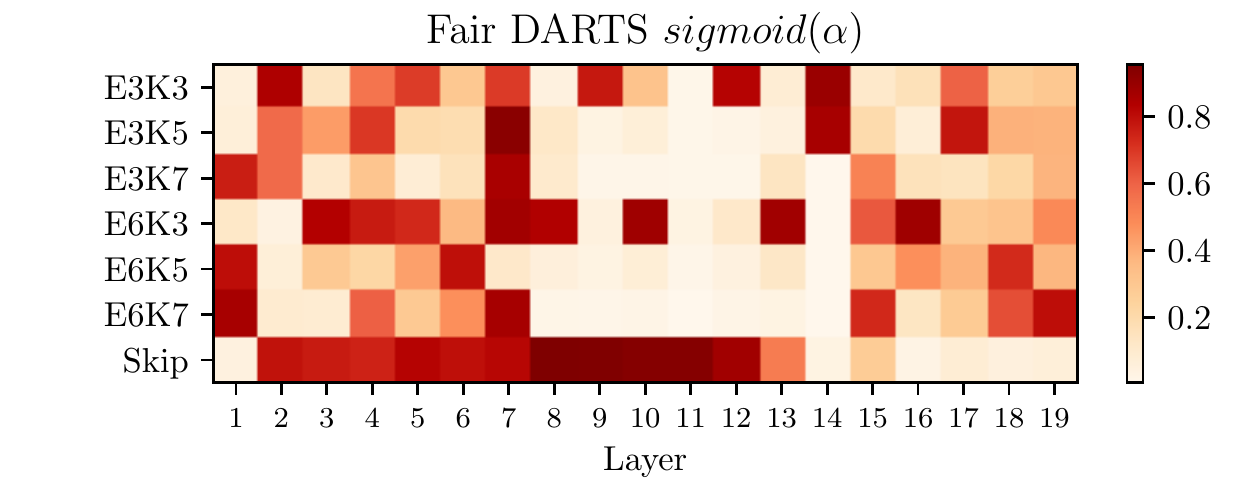}
	\caption{\textbf{Top:} Stacked area plot of the number of dominant operations\protect\footnotemark  of DARTS and Fair DARTS when searching on ImageNet in search space $S_2$ (19 searchable layers). \textbf{Bottom:} Heatmaps of softmax (DARTS) and sigmoid (Fair DARTS) values in the last searching epoch. DARTS finally chooses a shallow model (11 layers removed by activating skip connections only) which obtains $66.4\%$ top-1 accuracy. While in Fair DARTS, all operations develop independently that an excessive number of skip connections no longer leads to poor performance. Here it infers a deeper model (only one layer is removed) with $75.6\%$ top-1 accuracy} 
	\label{fig:num-skip-imagenet}
\end{figure}

In spite of these endeavors, the aggregation of skip connections in DARTS that noticed by  \cite{chen2019progressive,liang2019darts,bi2019stabilizing,zela2020understanding} has not been solved with perfection. Observing that the aggregation leads to a dramatic \textbf{performance collapse} for the resulting architecture, P-DARTS \cite{chen2019progressive} utilizes dropout as a workaround to restrict the number of skip connections during optimization. DARTS+ \cite{liang2019darts} directly puts a hard limit of two skip-connections per cell. RobustDARTS \cite{zela2020understanding} finds out that these solutions coincide with high validation loss curvatures. To some extent, these approaches consider the poor-performing models as impurities from the solution set, for which they intervene in the training process to filter them out.

On the contrary, we extend the solution set and revise the optimization process so that  aggregation of skip connections no longer causes the collapse.  
Moreover, there remains a \textbf{discrepancy problem} when discretizing continuous architecture encodings. DARTS \cite{liu2018darts} leaves it as future work, but till now it has not been deeply studied. We reiterate the basic premise of DARTS is that the continuous solution approximates a one-hot encoding. Intuitively, the smaller discrepancies are, the more consistent it will be when we transform a continuous solution back to a discrete one. We summarize our contributions as follows:
\footnotetext{In DARTS, it refers to the one with the highest architectural weight. In FairDARTS, it means the one whose $\sigma > \sigma_{threshold}$. Here we use $\sigma_{threshold}$ = 0.75. }

 \textbf{Firstly},  we disclose the root cause that leads to  the collapse of DARTS, which we later define as an \emph{unfair advantage} that drives skip connections into a monopoly state in \emph{exclusive competition}. These two indispensable factors work together to induce a performance collapse. Moreover, if either of the two conditions is broken, the collapse disappears.
 
\textbf{Secondly}, we propose the first \textbf{collaborative competition} approach by offering each operation an independent architectural weight. The unfair advantage no longer prevails as we break the second factor. Furthermore, to address the discrepancy between the continuous architecture encoding and the derived discrete one in our method, we propose a novel auxiliary loss, called \emph{zero-one loss}, to steer architectural weights towards their extremities, that is, either completely enabled or disabled. The discrepancy thus decreases to its minimum. 

\textbf{Thirdly},  based on the root cause of the collapse, we provide a unified perspective to view current DARTS cures for skip connections' aggregation. The majority of these works either make use of dropout \cite{srivastava2014dropout} on skip connections \cite{chen2019progressive,zela2020understanding}, or play with the later termed \emph{boundary epoch} by different early-stopping strategies \cite{liang2019darts,zela2020understanding}. They can all be regarded as preventing the first factor from taking effect.
Moreover, as a direct application, we can derive a hypothesis that adding Gaussian noise also disrupts the unfairness, which is later proved to be effective. 


\textbf{Lastly}, we conduct thorough experiments in two widely used  search spaces in both proxy and proxyless ways. Results show that our method can escape from performance collapse. We also achieve state-of-the-art networks on CIFAR-10 and ImageNet.

\section{Related Work}

Lately, neural architecture search \cite{zoph2017learning} has grown as a well-formed methodology to discover networks for various deep learning tasks. Endeavors have been made to reduce the enormous searching overhead with the weight-sharing mechanism \cite{brock2017smash,pham2018efficient,liu2018darts}. Especially in DARTS \cite{liu2018darts}, a nested gradient-descent algorithm is exploited to search for the graphical representation of architectures, which is born from gradient-based hyperparameter optimization \cite{maclaurin2015grad}.

Due to the limit of the DARTS search space,  ProxylessNAS   \cite{cai2018proxylessnas} and FBNet \cite{wu2018fbnet} apply DARTS in much larger search spaces based on MobileNetV2 \cite{sandler2018mobilenetv2}. ProxylessNAS also differs from DARTS in its supernet training process, where only two paths are activated, based on the assumption that one path is the best amongst all should be better than any single one. From a fairness point of view, as only two paths enhance their ability (get parameters updated) while others remain unchanged, it implicitly creates a bias. FBNet \cite{wu2018fbnet}, SNAS \cite{xie2018snas} and GDAS \cite{dong2019searching} utilize the differentiable Gumbel Softmax \cite{maddison2016concrete,jang2016categorical} to mimic one-hot encoding. However, the one-hot nature implies an exclusive competition,  which risks being exploited by unfair advantages.


Superficially, the most relevant work to ours is RobustDARTS \cite{zela2020understanding}. Under several simplified search spaces, they state that the found solutions generalize poorly when they coincide with high validation loss curvature, where the supernet with an excessive number of skip connections happens to be such a solution. Based on this observation, they impose early-stop regularization by tracking the largest eigenvalue. Instead, our method doesn't need to perform early stopping.


\section{The Downside of DARTS}
In this section, we aim to excavate the disadvantages of DARTS that possibly impede the searching performance. We first prepare a minimum background.

\subsection{Preliminary of Differentiable Architecture Search}
For the case of convolutional neural networks, DARTS \cite{liu2018darts} searches for a \emph{normal cell} and a \emph{reduction cell} to build up the final architecture. A cell is represented as a directed acyclic graph (DAG) of $N$ nodes in sequential order.  Each node stands for a feature map. The edge $e_{i,j}$ from node $i$ to $j$ operates on the input feature $x_i$ and its output is denoted as $o_{i,j}(x_i)$. The intermediate node $j$ gathers all inputs from the incoming edges,
\begin{align}
x_j = \sum_{i<j} o_{i, j} (x_i).
\end{align}

Let $\mathcal{O} = \{o_{i,j}^1, o_{i,j}^2, ..., o_{i,j}^M\}$ be the set of $M$ candidate operations on edge $e_{i,j}$. DARTS relaxes this categorical choice to a softmax  over all operations in $\mathcal{O}$ to form a mixed output:
\begin{align}{\label{eq:cifar_sum}}
\bar{o}_{i, j} (x) = \sum_{o \in \mathcal{O}} \frac{\exp(\alpha_{o_{i,j}})}{\sum_{o' \in \mathcal{O}} \exp(\alpha_{o'_{i,j}})} o(x),
\end{align}
where each operation $o_{i,j}$ is associated with a continuous coefficient $\alpha_{o_{i,j}}$. Regarding edge $e_{i,j}$, this softmax is utilized to approximate one-hot encoding $\beta_{i,j} = (\beta_{o_{i,j}^1}, \beta_{o_{i,j}^2}, ..., \beta_{o_{i,j}^M} )$. Formally, let $\alpha_{o_{i,j}}$ denote the architectural weights vector ($\alpha_{o_{i,j}^1}$, $\alpha_{o_{i,j}^2}$, ..., $\alpha_{o_{i,j}^M}$). DARTS thus assumes the following as a valid approximation,
\begin{equation}{\label{eq:softmax approx}}
	softmax(\alpha_{o_{i,j}})\approx \beta_{i,j}.
\end{equation}

The architecture search problem is reduced to learning $\alpha^*$ and network weights $w^*$ that minimize the validation loss $\mathcal{L}_{val} (w^*, \alpha^*)$. DARTS resolves this problem with a bi-level optimization, 
\begin{equation}\label{eq:darts-bi-level-opt}
\begin{split}
&\min_{\alpha} \mathcal{L}_{val} (w^*(\alpha), \alpha) \\
\text{ s.t. } &w^*(\alpha) = \text{argmin}_w  \mathcal{L}_{train} (w, \alpha).
\end{split}
\end{equation}

We also adopt two common search spaces, the DARTS \cite{liu2018darts} search space ($S_1$) and the ProxylessNAS \cite{cai2018proxylessnas} search space ($S_2$) with minor modifications. More details are given in Section \ref{sec:ss-supp} (supplementary).

In $S_2$, the output of the $l$-th layer is a softmax-weighted summation of $N$ choices. Formally, it can be written as
\begin{align}{\label{eq:imagenet_sum}}
	x_l = \sum_{k=1}^{N}\frac{\exp(\alpha_{l-1,l}^k)}{\sum_{j=1}^{N}\exp(\alpha_{l-1,l}^j)}o^k_{l-1,l}(x_{l-1}). 
\end{align} 

\subsection{Performance Collapse Caused by Intractable Skip Connections}{\label{sec:skip}}
DARTS suffers from significant performance decay when \emph{skip connections} become dominant \cite{chen2019progressive,liang2019darts}. It was described as a competition-and-cooperation issue in the bi-level optimization \cite{liang2019darts}. Still, the reason behind this behavior is not clear, we hereby provide a different perspective.

\begin{figure}[ht]
	\centering
	\includegraphics[scale=0.6]{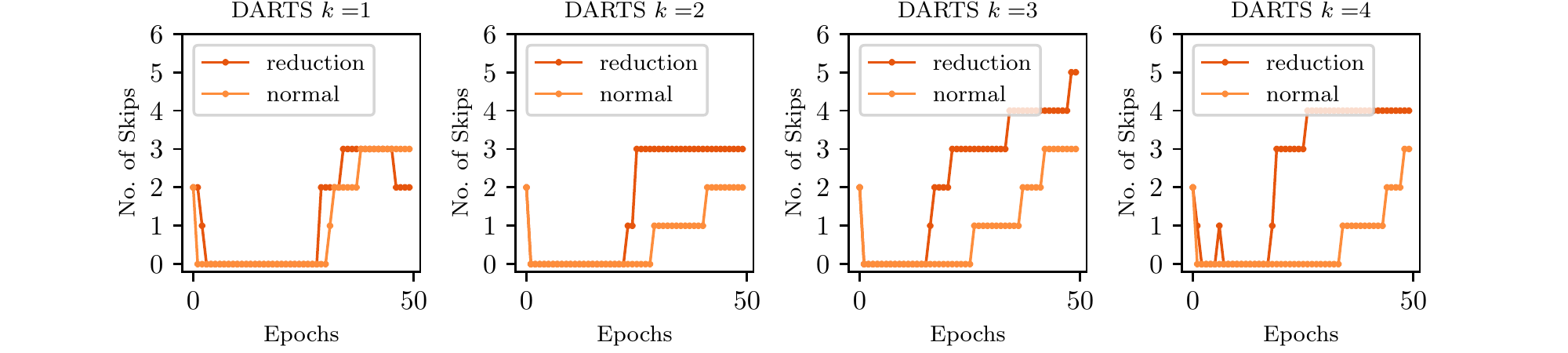}
	\caption{The number of dominant skip connections continues to grow when searching with DARTS (run $k=4$ times) on CIFAR-10 (in $S_1$)}
	\label{fig:num-skip-cifar}
\end{figure}

First, to confirm this issue, we run DARTS $k=4$ times with different random seeds. Following DARTS, we select 8 top-performing operations per cell (2 each for 4 intermediate nodes).  Here we say one operation is \emph{dominant} if it has top-2 $softmax(\alpha)$ among all incoming edges' candidates of a certain node. The results are shown in Fig.~\ref{fig:num-skip-cifar}. In the beginning, all operations are given the same opportunity. As the over-parameterized network gradually converges, there is an evident aggregation of skip connections after 20 epochs (5 out of 8 in an extreme case).

When we utilize DARTS directly on ImageNet in $S_2$, which is a single branch architecture, the same phenomenon rigorously reappears. The number of dominant skip-connections (highest $softmax(\alpha)$ among all operations in that layer) steadily increases and reaches 11 out of 19 layers  in the end, which is shown on the left of Fig.~\ref{fig:num-skip-imagenet}. 

But why is this happening? The underlying reasons are rarely discussed in depth. A brief and superficial analysis regarding information flow is given in \cite{chen2019progressive}. However, we claim that the reason for excessive skip connections is from \textbf{exclusive competition} among various operations. In Equation~\ref{eq:cifar_sum} and Equation~\ref{eq:imagenet_sum}, the skip connection is softmax-weighted and added to the output, which resembles a basic residual module as in ResNet \cite{he2016deep}. While this module greatly benefits the training, the architectural weight of a skip connection increases much faster than its competitors. Moreover,  \emph{the softmax operation inherently provides an exclusive competition since increasing one is at the cost of suppressing others}. As a result, skip connections become gradually dominant during optimization.  
We have to keep in mind that skip connection works well because it is in cooperation with convolutions \cite{he2016deep}. 
However, DARTS picks the top-performing one (skip connection here) and discards its collaborator (convolution), which results in a degenerate model.

\begin{figure}[ht]
	\centering
	\includegraphics[scale=0.5]{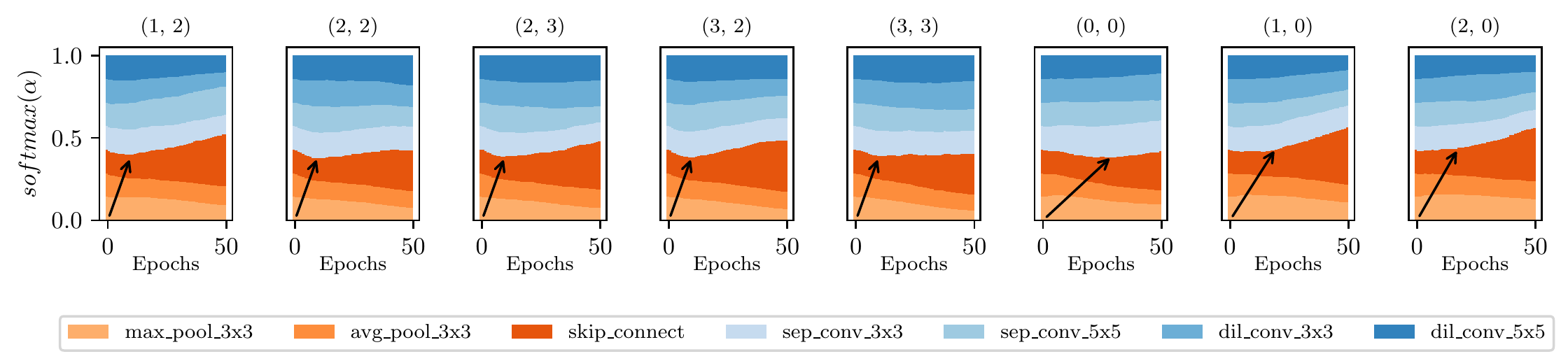}
	\caption{The softmax evolution where skip connections gradually become dominant when running DARTS on CIFAR-10 (in $S_1$).  Last two subplots of edge (1,0) and (2,0) are from the normal cell, the rest are from the reduction cell. Black arrows point to boundary epochs where skip connections start to demonstrate its strength}
	\label{fig:darts-bar-skip-dominant-edge}
\end{figure}

We further study this effect from the experiments on CIFAR-10 by recording the competition progress in Fig.~\ref{fig:darts-bar-skip-dominant-edge}. The derived model has 8 skip connections in total\footnote{corresponding to the experiment ($k=3$ ) in Fig.~\ref{fig:num-skip-cifar}.}. ResNet \cite{he2016deep} discovers that \emph{skip connections begin to demonstrate power after a few epochs compared with models without them}. Interestingly, a similar phenomenon  is also observed in our experiments. We term this tipping point a \emph{boundary epoch}. The boundary epochs may vary from edge to edge, but are generally at the early stage. From Fig.~\ref{fig:darts-bar-skip-dominant-edge}, we observe that skip connections colored in red-orange progressively obtain higher architectural weights after some certain boundary epochs. Meantime, other operations are suppressed and steadily decline. We consider this benefit from the residual module as an unfair advantage by Definition \ref{def:unfair}.

\begin{definition}
	\label{def:unfair}
	\textbf{Unfair Advantage.}  Suppose that choosing one operation among others is a competition. This competition is deemed \textbf{exclusive} when only restricted operations  can be selected. An operation in an exclusive competition is said to have an \textbf{unfair advantage} if this advantage contributes more to competition than to the performance of a resulted network.
\end{definition}

From the above discussion, we can draw 
\textbf{Insight 1: The root cause of excessive skip connections is the inherent unfair competition.} The skip connection has an unfair advantage by forming a residual module which is convenient for the supernet training, but not equally beneficial for the performance of the outcome network where the residual module is broken.


\subsection{Non-negligible Discrepancy of Discretization}\label{sec:problem of discrepancy}
\begin{figure}[ht]
	\centering
	\includegraphics[width=0.4\textwidth,scale=0.6]{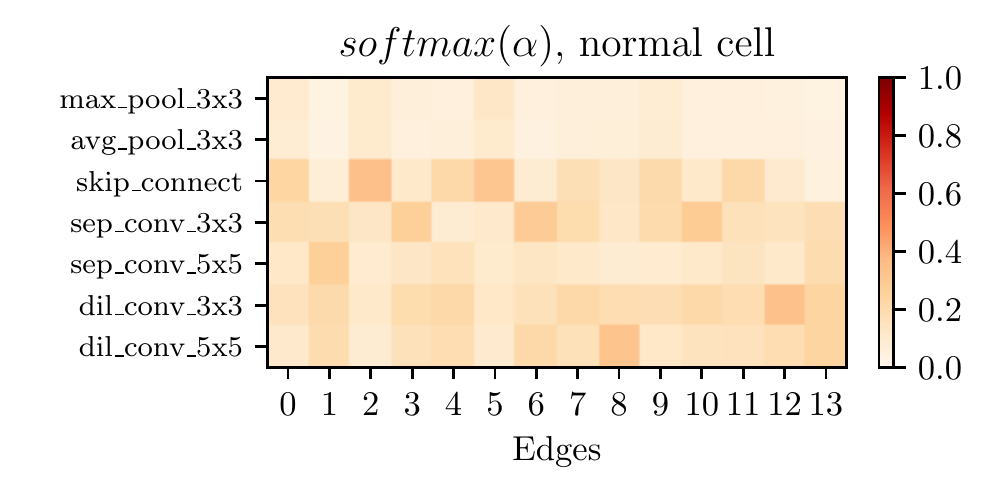}
	\includegraphics[width=0.4\textwidth,scale=0.6]{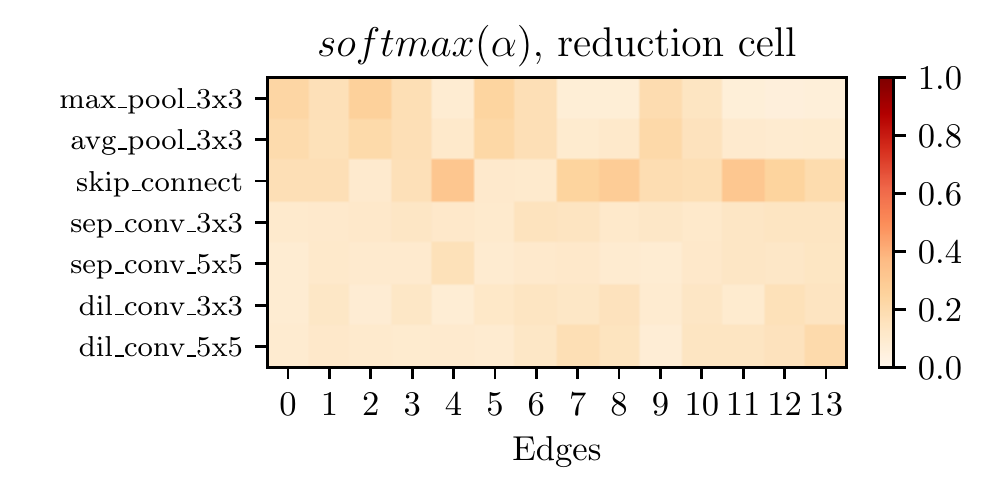}
	\caption{Heatmap of softmax values in the normal cell and the reduction cell at the last searching epoch when running DARTS on CIFAR-10 (in search space $S_1$)}
	\label{fig:alpha-heatmap-darts-cifar-imagenet}
\end{figure}

Apart from the above issue, DARTS reports that it suffers from discrepancies when discretizing continuous encodings \cite{liu2018darts}. 
To verify the problem, we run DARTS in $S_1$ on CIFAR-10, and in $S_2$ on ImageNet. The values of $softmax(\alpha)$ of the last iteration are displayed in Fig.~\ref{fig:alpha-heatmap-darts-cifar-imagenet} ($S_1$) and on the bottom left of Fig.~\ref{fig:num-skip-imagenet} ($S_2$). For $S_1$,  the largest value is about 0.3 while the smallest one is above 0.1\footnote{We run DARTS 4 times and it holds every time.}. This range is somewhat too narrow to differentiate `good' operations from `bad'. For instance on edge 2 of the reduction cell, the values are very close to each other, [0.174, 0.170, 0.176, 0.112, 0.116, 0.132, 0.118], it's hard to say that an operation weighted by 0.176 is better than the other by 0.174.  
For $S_2$, the top-1 values are not so evidently particular from layer 2 to 7. Take the second layer for example, we have to use [0.235, 0.057, 0.17,  0.016, 0.187, 0.269, 0.066] to approximate [0, 0, 0, 0, 0, 1, 0]. This again confirms the existence of discrepancy.


In summary, DARTS is usually far from a good resemblance to a one-hot representation as required by its premise in Equation~\ref{eq:softmax approx}. We often have to make ambiguous choices without high confidence. Hence, we learn 
\textbf{Insight 2: Relaxing from discrete categorical choices to continuous ones should make a close approximation}.

\section{Fair DARTS}
\subsection{Stepping out the Pitfalls of Skip Connections}

Based on \textbf{Insight 1}, we propose a \emph{cooperative mechanism} to eliminate the existing unfair advantage. Not only should we exploit skip connection for smoother information flow, but we also have to provide equal opportunities for other operations. In a word, they need to avoid being trapped by unfair advantage from skip connections. On this regard, we apply a \emph{sigmoid activation} ($\sigma$) for each $\alpha_{i,j}$, so that each operation can be switched on or off independently without being suppressed. Formally, we replace Equation~\ref{eq:cifar_sum} with the following,

\begin{align}{\label{eq:fairsum}}
\bar{o}_{i, j} (x) = \sum_{o \in \mathcal{O}} \sigma(\alpha_{o_{i,j}})  o(x).
\end{align}

It's trivial to show that even if $\sigma(\alpha_{skip})$ saturates to 1, other operations still can be optimized cooperatively. Promising operations continue to grow their architectural weights to reduce $\mathcal{L}_{val}$, which leads to a \textbf{multi-hot} approximation. Instead, DARTS attempts to derive a one-hot estimation. The difference is that we have extended the solution set. Consequently, it allows us to tackle the discretization discrepancy. We are left to find out how to drive $\sigma(\alpha)$ towards each extremity (0 or 1). Next, we discuss it in greater detail.

\subsection{Resolve Discrepancy from Continuous Representation to Discrete Encoding}{\label{sec:loss}}
To abide by \textbf{Insight 2}, we explicitly coerce an extra loss called \emph{zero-one loss} to push the sigmoid value of architectural weights towards 0 or 1. Let $L_{0-1} = f(z)$ denote this loss component, where $z=\sigma(\alpha)$.  To achieve our goal, the loss design must meet three basic criteria, 
\begin{enumerate*}
\item It needs to have a global maximum at $z=0.5$ (a fair starting point) and a global minimum at 0 and 1.
\item The gradient magnitude $\frac{df}{dz}|_{z\approx0.5}$ has to be adequately small to allow architectural weights to fluctuate, but large enough to attract $z$ towards 0 or 1 when they are a bit far from 0.5.
\item It should be differentiable for backpropagation. 
\end{enumerate*}

According to the first requirement, we move $\sigma(\alpha)$ away from 0.5 towards 0 or 1 to minimize the discretization gap. The second one enacts explicit necessary constraints. Particularly, small gradients around the peak avoid stepping easily into two ends. Larger gradients around 0 and 1 instead help to quickly capture $z$ nearby. Quite straightforward, we come up with a loss function to meet the above requirements, formally as,
\begin{equation}
	L_{0-1} = -\frac{1}{N}\sum_{i}^{N}(\sigma(\alpha_i)-0.5)^2
\end{equation}
In order to control its strength, we weight this loss by a coefficient $w_{0-1}$, thus the total loss for $\alpha$ is formulated as, 
\begin{equation}{\label{eq:total_loss}}
	L_{total} = \mathcal{L}_{val} (w^*(\alpha), \alpha) + w_{0-1}L_{0-1}.
\end{equation}

Like DARTS \cite{liu2018darts}, the architectural weights can be optimized through backpropagation. From Equation~\ref{eq:total_loss}, the search objective is to find an architecture of high accuracy  with a good approximation from a continuous encoding to a discrete one.

Moreover, the second requirement is indispensable, otherwise the gradient-based approach may step into local minimum too early. Here we design another loss as a negative example. Let $L_{0-1}'$ be the following,

\begin{equation}
L_{0-1}' = -\frac{1}{N}\sum_{i}^{N}|(\sigma(\alpha_i)-0.5)|.
\end{equation}

\begin{figure}[ht]
	\centering
	\includegraphics[scale=0.45]{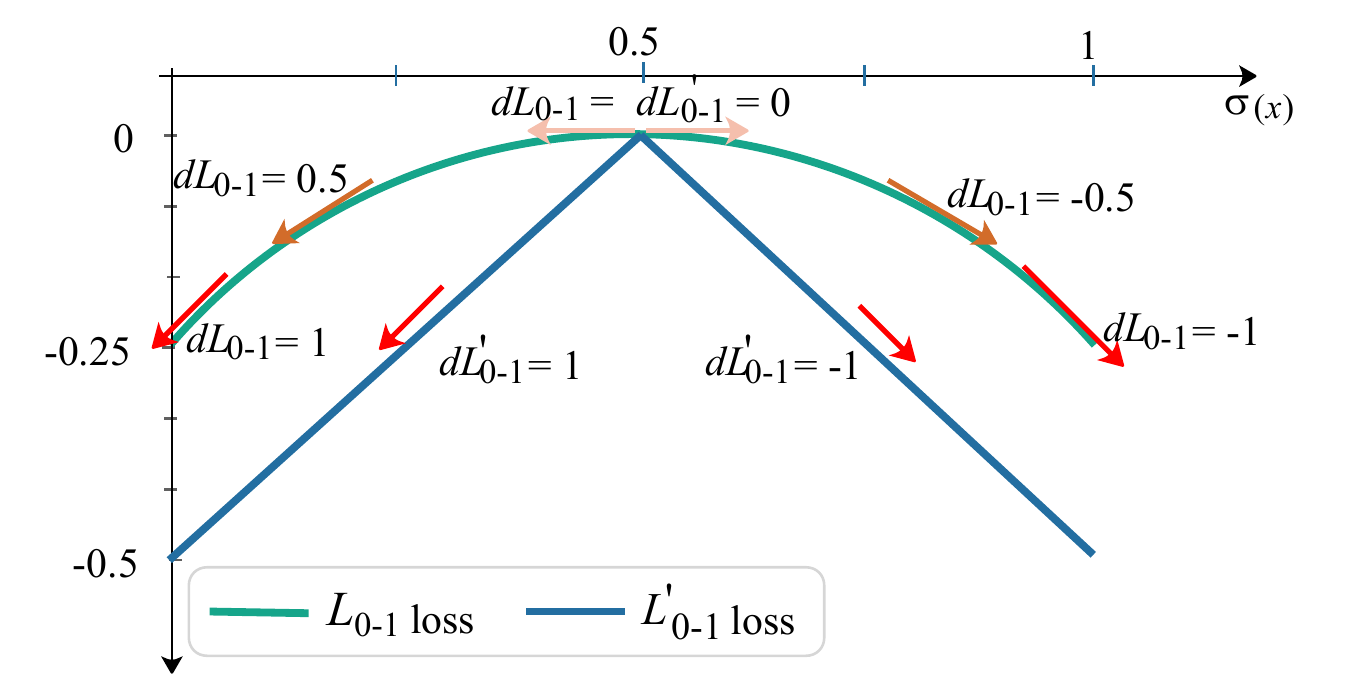}
	\caption{Illustration about the auxiliary loss design: $L_{0-1}$  (proposed) and $L'_{0-1}$ (control)}
	\label{fig:01los}
\end{figure}

It's trivial to see that $\frac{d|z-0.5|}{dz}|_{z>0.5} = 1 $ and $\frac{d|z-0.5|}{dz}|_{z<0.5} = -1 $. Once $z$ stays away from 0.5, it may receive the same gradient (1 or -1) in the later iterations, thus rapidly pushing the architectural weights towards two ends. This phenomenon is illustrated in Fig.~\ref{fig:01los}.

To conclude, by combining Equation~\ref{eq:darts-bi-level-opt}, \ref{eq:fairsum} and \ref{eq:total_loss}, our method which we call Fair DARTS, can be now formally written as
\begin{equation}\label{eq:fairidarts-bi-level-opt}
\begin{split}
&\min_{\alpha} \mathcal{L}_{val} (w^*(\alpha), \alpha) + w_{0-1}L_{0-1} \\
\text{ s.t. } &w^*(\alpha) = \text{argmin}_w  \mathcal{L}_{train} (w, \alpha). \\
&\bar{o}_{i, j} (x)= \sum_{o \in \mathcal{O}} \sigma(\alpha_{o_{i,j}})  o(x).
\end{split}
\end{equation}

It is also important to recognize that our zero-one loss is specially designed for Fair DARTS. Pushing $\sigma(\alpha)$ of one edge towards 0 or 1 is independent of others. It cannot be directly applied to DARTS given the exclusive competition by softmax.  
As the architectural weights converge to their extremities, it's natural to use a threshold value $\sigma_{threshold}$ in our approach to infer submodels instead of \emph{argmax} .

\section{Experiments and Results}

\subsection{Searching Architectures for CIFAR-10}

At the search stage, we use similar hyperparameters and tricks as \cite{liu2018darts}. We apply the \emph{first-order} optimization and it takes 10 GPU hours. All experiments are done on a Tesla V100.
We select our target models with $\sigma_{threshold} = 0.85$\footnote{The maximum number of edges for a node is also limited to 2 as in DARTS.}. We use the same data processing and training trick as \cite{liu2018darts,chen2019progressive}. 

Our collaborative approach performs well with skip connections aggregation. To verify this, we repetitively search 7 times on different random seeds and report the number of skip connections in Fig.~\ref{fig:num-skip-cifar-fair-darts} (see supplementary). Since the number of skip connections is more reasonable, we obtain an average top-1 accuracy 97.46$\%$. Especially, the smallest FairDARTS-a reaches $97.46\%$ accuracy on CIFAR-10 with reduced parameters and multiply-adds. A complete result of FairDARTS searched cells are shown in the supplementary (Fig.~\ref{fig:normal-reduce-architecture}, Fig.~\ref{fig:fairdarts-b-normal-reduce-architecture} and Table~\ref{tab:fairdarts_models}).

\setlength{\tabcolsep}{4pt}
\begin{table}
	\begin{center}
		\caption{Comparison of architectures on CIFAR-10. $^\dagger$: MultAdds computed using the genotypes provided by the authors. $^\star$: Averaged on training the best model for several times .   $^\ddagger$: Averaged on models from 7 runs of FairDARTS (Search + Full Train)} 
		\label{tab:comparison-cifar10}
	  \begin{footnotesize}
		\begin{tabular}{*{6}{l}} 			
		\hline\noalign{\smallskip}
			Models  & Params (M) & $\times+$ (M) & Top-1 (\%) & Type  \\
			\hline\noalign{\smallskip}
			NASNet-A \cite{zoph2017learning}  & 3.3 & 608$^\dagger$  &  97.35 & RL \\
			ENAS \cite{pham2018efficient} & 4.6 & 626$^\dagger$ & 97.11 & RL     \\	
			MdeNAS\cite{zheng2019multinomial} & 3.6 & 599$^\dagger$ & 97.45 & MDL \\
			\hline\noalign{\smallskip}
			DARTS(second order)$^\star$\cite{liu2018darts} & 3.3 & 528$^\dagger$ & 97.24$\pm0.09$ & GD \\ 
			SNAS$^\star$ \cite{xie2018snas}  & 2.8 & 422$^\dagger$ & 97.15$\pm0.02$ & GD\\
			GDAS \cite{dong2019searching} & 3.37 & 519$^\dagger$ & 97.07 &GD \\
			SGAS (Cri.2 avg.) \cite{li2019sgas} & 3.9$\pm$0.22$^\dagger$ & 640$\pm$39$^\dagger$ & 97.33$\pm0.21$ & GD\\
			P-DARTS \cite{chen2019progressive} & 3.4 & 532$^\dagger$ & 97.5 & GD \\ 
			PC-DARTS \cite{xu2019pc} & 3.6 & 558$^\dagger$ & 97.43 & GD \\ 
			RDARTS \cite{zela2020understanding} &-& -& 97.05 & GD \\
			FairDARTS-a & \textbf{2.8} & \textbf{373} & 97.46 & GD \\ 
			FairDARTS$^\ddagger$ & 3.32$\pm$0.46  & 458$\pm$61 & 97.46$\pm0.05$ & GD \\
			\hline
		\end{tabular}
		\end{footnotesize}
	\end{center}
\end{table}\setlength{\tabcolsep}{1.4pt}

\subsection{Transferring to ImageNet}
As a common practice, we transfer two searched cells (FairDARTS-a and b\footnote{Their architectures are given in Fig.~\ref{fig:normal-reduce-architecture} and ~\ref{fig:fairdarts-b-normal-reduce-architecture} (supplementary).}) to  ImageNet. We keep the same configurations and use the identical training tricks as DARTS \cite{liu2018darts}. Compared with SNAS \cite{xie2018snas} and DARTS, FairDARTS-A only uses 3.6M number of parameters and 417M multiply-adds to obtain $73.7\%$ top-1 accuracy on ImageNet validation set. FairDARTS-B also achieves state-of-the-art $75.1\%$ in $S_1$ with a smaller number of parameters than comparable counterparts. 


\setlength{\tabcolsep}{4pt}
\begin{table}
	\begin{center}
	\caption{Comparison of architectures on ImageNet. $^\star$: Based on its published code. $^\dagger$: Searched on CIFAR-10. $^{\dagger\dagger}$: Searched on CIFAR-100. $^\ddagger$: Searched on ImageNet (cost more than those transferred). $^\bullet$: in GPU days. $^{\diamond}$: w/ SE and Swish} 
	\label{tab:comparison-imagenet}
	  \begin{footnotesize}
		\begin{tabular}{*{7}{l}} 			
		\hline\noalign{\smallskip}
			Models & $\times+$ (M)  &Params (M) & Top-1 (\%) & Top-5 (\%) & Cost$^\bullet$ \\
			\hline\noalign{\smallskip} 
			MobileNetV2(1.4) \cite{sandler2018mobilenetv2}   & 585 & 6.9 & 74.7 & 92.2 &-\\
			\hline\noalign{\smallskip}
			NASNet-A \cite{zoph2017learning}  & 564 & 5.3 &74.0 & 91.6& 2000\\
			AmoebaNet-A\cite{real2018regularized} & 555  & 5.1 & 74.5 &92.0& 3150 \\
                       MnasNet-92 \cite{tan2018mnasnet}  & 388 & 3.9 & 74.79 & 92.1 & 1667\\ 	
		    \hline\noalign{\smallskip}
			DARTS \cite{liu2018darts} & 574 & 4.7 & 73.3 & 91.3& 4\\
			FBNet-C \cite{wu2018fbnet}   & 375 & 5.5 &  74.9 & 92.3 & 9 \\ 
			Proxyless GPU$^\ddagger$ \cite{cai2018proxylessnas}  & 465$^*$ & 7.1  & 75.1 & 92.4 & 8.3\\
			FairNAS-C$^\ddagger$ \cite{chu2019fairnas} &321 & 4.4 & 74.7 &92.1 & 10 \\
			SNAS \cite{xie2018snas}  &522&4.3&72.7&90.8 & 1.5\\
			GDAS \cite{dong2019searching}  &581&5.3&74.0&91.5 & 0.2\\
			P-DARTS$^{\dagger\dagger} $\cite{chen2019progressive}& 577 & 5.1 & 74.9$^{*}$ & 92.3$^{*}$ & 0.3 \\  
			PC-DARTS$^\dagger$ \cite{xu2019pc} & 586 & 5.3 & 74.9 & 92.2& 3.8\\ 
			FairDARTS-A$^\dagger$ & 417 & 3.6 &73.7 & 91.7& 0.4 \\
		        FairDARTS-B$^\dagger$ & 541 & 4.8 &75.1 & 92.5& 0.4 \\
			FairDARTS-C$^\ddagger$ & 380 &4.2 &75.1 & 92.4 &3\\
			\textbf{FairDARTS-D}$^\ddagger$ & 440 & 4.3 & \textbf{75.6} & \textbf{92.6}&3 \\
			\hline\noalign{\smallskip}
			MobileNetV3 \cite{howard2019searching} & 219 & 5.4 &75.2 & 92.2 & -\\
			MoGA-A \cite{chu2019moga} & 304 & 5.1 & 75.9 & 92.8 & 12 \\
			MixNet-M \cite{tan2020mixconv} &360 & 5.0 & 77.0 & 93.3& -\\
			EfficientNet B0 \cite{tan2019efficientnet} &390 & 5.3  & 76.3 & 93.2 & - \\
			SCARLET-A \cite{chu2019scarletnas} & 365 & 6.7 &   76.9 &93.4 & 10 \\
			\textbf{FairDARTS-C}$^\diamond$ & 386 & 5.3 & \textbf{77.2} & \textbf{93.5} & 3 \\

			\hline\noalign{\smallskip}
		\end{tabular}
		\end{footnotesize}
	\end{center}
\end{table}\setlength{\tabcolsep}{1.4pt}

\subsection{Searching Proxylessly on ImageNet}{\label{sec:proxyless}}

Relaxing exclusive competition to collaboration greatly extends the size of the search space. In ProxylessNAS \cite{cai2018proxylessnas}, there are 19 searchable layers and each layer contains 7 choices, consisting of $7^{19}$ possible models. In our approach, every choice can be activated independently, thus, $S_2$ contains $({2^{7}})^{19}=128^{19}$ possible models. To our knowledge, this is a most gigantic search space ever proposed, about $18^{19}$ times that of \cite{cai2018proxylessnas}. 

For this search phase, we train for 30 epochs with a batch size of 1024, which takes about 3 GPU days. The final architectural weight matrix (after sigmoid activation) on the bottom right of Fig.~\ref{fig:num-skip-imagenet} is used to derive target models. Under this cooperative setting, the skip connections and other inverted bottleneck blocks can be both chosen to work together, where the former facilitates the training and the latter learn the residual information\cite{li2017convergence}. In contrast, under the competitive setting of DARTS, it's impossible to achieve this, as shown in the bottom left of Fig.~\ref{fig:num-skip-imagenet}. Within 19 layers have 11 skip connection operation is preferred, which cuts down the overall depth of searchable layers to 8.

\begin{figure*}[ht]
	\centering
	\includegraphics[scale=0.26]{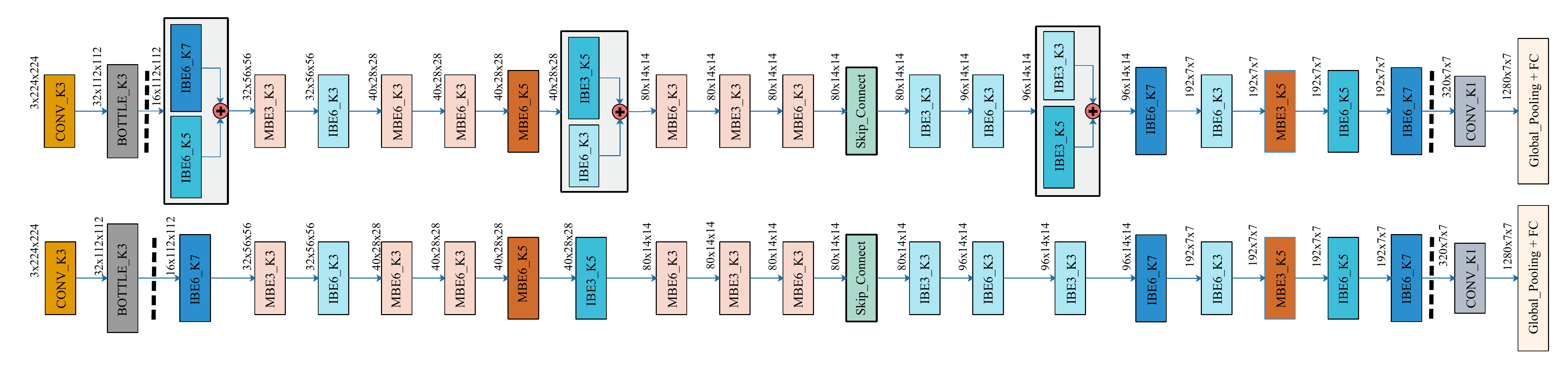}
	\caption{The Architecture of Fair DARTS-D (top) and C (bottom). IBE$x$\_K$y$ refers to an inverted bottleneck without an inset skip connection, while MBE$x$\_K$y$ is the one with it. BOTTLE\_K3 is the inverted bottleneck without expansion}
	\label{fig:fair-darts-arch}
\end{figure*}

To be fair, we select at most two choices per layer if there are more than two  above $\sigma_{threshold}$ (0.75) and use the same training tricks as \cite{tan2018mnasnet}. We exclude squeeze and excitation \cite{hu2018squeeze} and refrain from using AutoAugment \cite{cubuk2018autoaugment} tricks though they can boost the classification accuracy further. The searched model FairDARTS-D is shown in Fig.~\ref{fig:fair-darts-arch}, which places the summation of two inverted bottleneck blocks nearby the down-sampling stage to keep more information. It also utilizes large kernels and big expansion blocks at the tail end. Further, We raise the $\sigma_{threshold}$ as 0.8 to get a more lightweight model FairDARTS-C. FairDARTS-C achieves $75.1\%$ top-1 accuracy using only 4.2 M number of parameters. To make comparisons with EfficientNetB0 \cite{tan2019efficientnet}, MobileNetV3 \cite{howard2019searching} and MixNet \cite{tan2020mixconv}, FairDARTS-C obtains $77.2\%$ top-1 accuracy with the same tricks such as squeeze-and-excitation \cite{hu2018squeeze}, AutoAugment \cite{cubuk2018autoaugment} and Swish \cite{ramachandran2017searching}.

\section{Ablation Study and Analysis}

\subsection{Removing Skip Connections from $S_1$}
\begin{figure}[ht]
	\centering
	\includegraphics[scale=0.53]{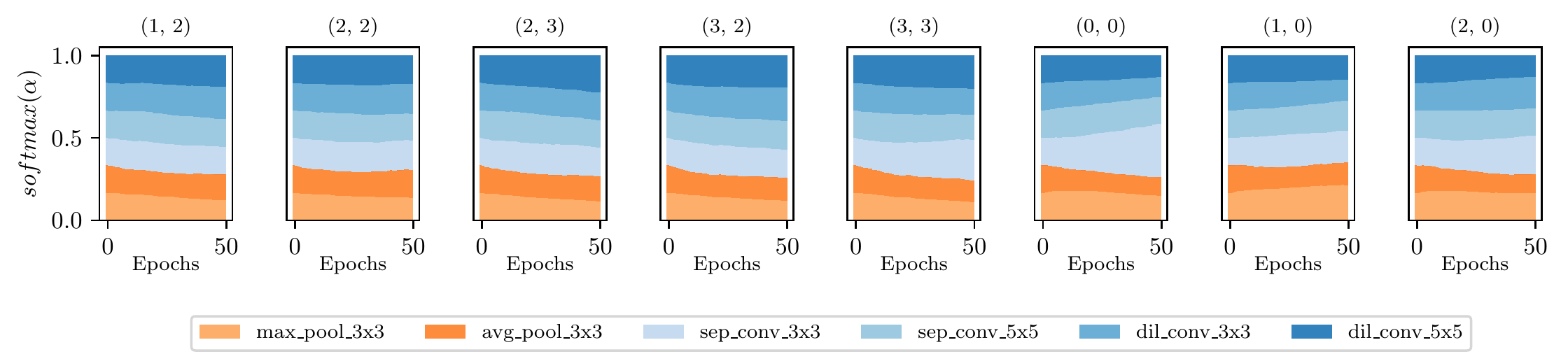}
	\caption{Stacked area plot of the softmax evolution in  $S_1\setminus\{skip\}$ when running DARTS on CIFAR-10. With unfair advantages removed, all operations enjoy a fair treatment}
	\label{fig:darts-bar-no-skip-dominant-edge}
\end{figure}

As unfair advantages are mainly from skip connections, if we remove them from $S_1$ and get the reduced search space $S_1\setminus\{skip\}$, we should expect a fair play even in an exclusive competition. Several runs of this experiment also show that there is indeed no more prevailing operations that suppress others, including other parameter-less ones like max-pooling and average pooling (Fig.~\ref{fig:darts-bar-no-skip-dominant-edge}). For $S_1\setminus\{skip\}$, we run all the experiments with 7 different random seeds and we train the searched models from scratch. The best models ($96.88\pm0.18\%$) are slightly higher than  DARTS ($96.76\pm0.32\%$)\footnote{This differs from DARTS' reported values as it trains one model for several times.}, but lower than FairDARTS ($97.41\pm0.14\%$) in  $S_1$. The lowered accuracy indicates that adequate  skip connections are indeed beneficial for accuracy.

\subsection{How Does Zero-One Loss Matter?}\label{sec:zero-one-loss}

\begin{figure}[ht]
	\centering
	\includegraphics[scale=0.5]{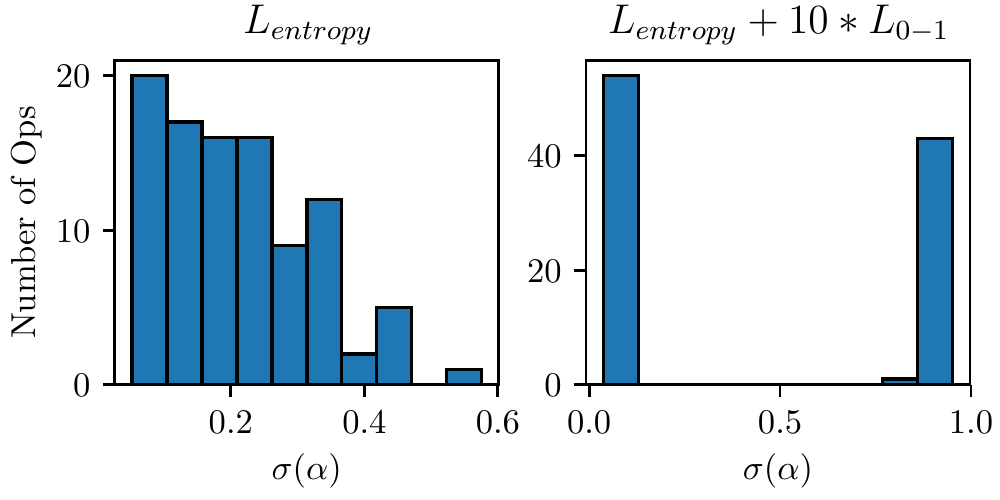}
	\hskip 0.1in
	\includegraphics[scale=0.45]{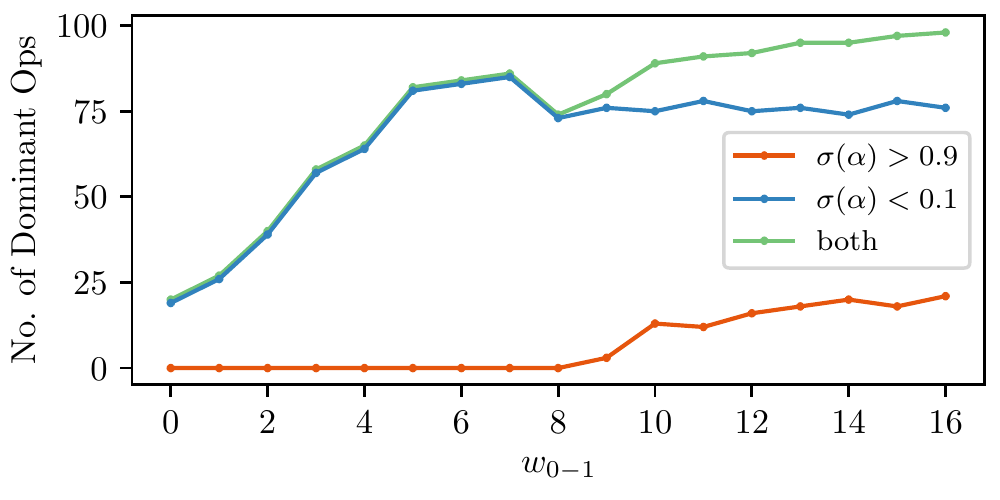}
	\caption{\textbf{Left:} Histogram of sigmoid values in the last searching epoch without (left) and with $L_{0-1}$ (right). On the right, this auxiliary loss has pushed the values towards 0 or 1. \textbf{Right:} Number of dominant operations in the last searching epoch running Fair DARTS on CIFAR-10 w.r.t the sensitivity weight $w_{0-1}$ of auxiliary loss $L_{0-1}$}
	\label{fig:histo-loss-and-weight-dominant}
\end{figure}

\textbf{Removing Zero-One Loss.} We design two comparison groups for Fair DARTS with and without \emph{zero-one loss}. Other settings are kept the same. We count the distribution of the sigmoid outputs from architectural weights and plot it on the left of Fig.~\ref{fig:histo-loss-and-weight-dominant}.  The one without zero-one loss covers a wide range between 0 and 0.6. So we have to make ambiguous choices again. Whereas the proposed loss has narrowed the distribution into two ends around 0 and 1. To further evaluate the influences of removing $L_{0-1}$, we repeat the searching for 7 times using different random seeds. The averaged top-1 accuracy for these models is 97.33$\pm$0.15 (532M FLOPS on average, 74 M more than FairDARTS with $L_{0-1}$). 
Therefore, although the unfair advantage is balanced, making ambiguous choices still bring noises to the final search result, which is better solved by $L_{0-1}$. Discrepancy elimination seems to helps find more light-weight and accurate models.

\textbf{Zero-One Loss Design.}
We run two experiments on CIFAR-10, one with $L_{0-1}$ (proposed) and the other $L'_{0-1}$ (control). To some extent, $L_{0-1}$ allows stepping out of the local minimum while $L_{0-1}'$ selects operations only at an early stage which depends greatly on the initialization. This matches our analysis in Section~\ref{sec:loss}. The detailed results under both loss functions are shown in Fig.~\ref{fig:l1-l2-sigmoid-comparison} (supplementary).

\textbf{Loss Sensitivity.} As the weight $w_{0-1}$ of this auxiliary loss goes higher, it should squeeze more operations towards two ends, but it must not overshadow the main entropy loss. We perform several experiments where an integer $w_{0-1}$ varies within [0, 16]. The right of Fig.~\ref{fig:histo-loss-and-weight-dominant} shows the final number of dominant operations for each. We select a reasonable $w_{0-1}=10$ for the best trade-off.




%

\subsection{Discussions From Fairness Perspective}\label{sec:discussion}

We review the existing methods that seek to avoid the discussed weaknesses. In general, adding dropouts  \cite{chen2019progressive,zela2020understanding} to operations is similar to blending them with a simple additive Gaussian noise, both reduce the performance gain from unfair advantages. Early-stopping \cite{liang2019darts} avoids the case before unfairness prevails.


\textbf{Adding dropout to skip connections reduces unfairness.}
The operation-level dropout \cite{srivastava2014dropout} inserted after skip connections by P-DARTS \cite{chen2019progressive} can be viewed as an alleviation of unfair advantage.  However, it comes with two obvious drawbacks. First, this dropout rate is hard to tune. Second, it is not so effective that they must involve another prior: setting the number of skip connections in the final cell to $M$. This is a very strong prior for searching good architectures \cite{liang2019darts}. 

\textbf{Adding dropout to all operations also helps}. Dropout troubles the training of skip connections and thus weakens the unfair advantage. Reasonably, higher dropout rates are more effective, especially for parameter-free operations. Therefore, RobustDARTS \cite{zela2020understanding} adds dropout to all operations and obtains promising results.


\textbf{Early stopping matters.}
DARTS+ \cite{liang2019darts} explicitly limits the maximum number of skip connections, which can be viewed as an early-stopping strategy nearby the previously mentioned \emph{boundary epoch}, right before too many skip connections rise into power. RobustDARTS \cite{zela2020understanding} also exploits early-stopping when the maximal Hessian eigenvalues change too fast.

\setlength{\tabcolsep}{4pt}
\begin{table}
	\begin{center}
	\caption{\textbf{Experiment 1:} Random sampling (7 models each, averaged) from regularized search space ($M$ = 2). \textbf{Experiment 2:} Adding Gaussian noise to DARTS (repeated 4 times, averaged)}
	\label{tab:noisy-darts}	
		\begin{footnotesize}
		\begin{tabular}{lc}
				\hline\noalign{\smallskip}
				Methods  & CIFAR-10 Top-1 Acc (\%) \\
				\hline\noalign{\smallskip} 
				Random (M=2)  & 97.01 $\pm0.24 $  \\
				Random (M=2, MultAdds $\ge$ 500M) $ $  & 97.14 $\pm 0.28$  \\
				\hline\noalign{\smallskip}
				DARTS + Gaussian (cosine decay)  & 97.12 $\pm 0.23$  \\

				\hline
			\end{tabular}
		\end{footnotesize}
	\end{center}
\end{table}

\textbf{Limiting the number of skip connections is a strong prior.}  In the regularized search space of P-DARTS \cite{chen2019progressive}  and DARTS+ \cite{liang2019darts}, we find that simply by restricting $M=2$, it is possible to generate competitive models even \emph{without searching}. We randomly sample models from their search space and report the results in Table~\ref{tab:noisy-darts}. In Experiment 1, the second group restricts the multiply-adds to be above 500M, to further leverage the average performance.  Surprisingly, both groups outperform DARTS \cite{liu2018darts}.  


\textbf{Random noise can break unfair advantage.} Based on our theory, we can boldly postulate that adding a random noise also disrupts the unfair advantage. Therefore, on top of DARTS \cite{liu2018darts}, we mix the skip connections' architectural weights with a standard Gaussian noise $\mathcal{N}(0,1)$, which has a cosine decay on 50 epochs. The results strongly confirm our hypothesis, as shown in Table~\ref{tab:noisy-darts}. We repeat it 4 times to have similar results. 

\textbf{Remove unfair advantages or destroy the exclusive competition?} In principle, we can break either one of the indispensable factors to avoid collapse. However,  FairDARTS breaks the latter which is simple and effective. Besides, it paves the way to eliminate the discrepancy by scheming an auxiliary loss $L_{0-1}$. Otherwise, the discrepancy issue remains hard to solve. However, to tackle the discrepancy issue, it's promising that the existing approaches might benefit from tricks like $L_{0-1}$. This remains to be our future work.  

\section{Conclusion}

We unveil two indispensable factors of the DARTS's aggregation of excessive skip connections: \textbf{unfair advantages} and \textbf{exclusive competition}. We prove that breaking any one of them can improve the robustness. First, by allowing collaborative competition, each operation develops its architectural weight independently. Meanwhile, the non-negligible discrepancy of discretization is reduced at maximum by coercing a novel auxiliary loss which polarizes the architectural weights. In this regard, we achieve state-of-the-art performance both on CIFAR-10 and ImageNet. Second, disturbing the differentiable process with a Gaussian noise removes unfair advantage which leads to competitive results. 
 
One of our future work is to make it more memory-friendly. As Gumbel softmax is used to replace categorical distribution \cite{wu2018fbnet}, is there a similar way to our approach? More methods remain to be explored on our basis.

\clearpage
\title{Supplementary of  ``Fair DARTS: Eliminating Unfair Advantages in Differentiable Architecture Search"}


\titlerunning{Fair DARTS (Supplementary)}
\author{Xiangxiang Chu\inst{1}\orcidID{0000-0003-2548-0605}  \and
	Tianbao Zhou\inst{2}\orcidID{0000-0002-2133-059X}  \and
	Bo Zhang\inst{1}\orcidID{0000-0003-0564-617X}  \and
	Jixiang Li\inst{1}\orcidID{0000-0001-5949-1498}}
\authorrunning{Chu et al.}
%
\institute{Xiaomi AI Lab \\ \email{\{chuxiangxiang,zhangbo11,lijixiang\}@xiaomi.com} \and Minzu University of China \\\email{tianbaochou@163.com}}

\maketitle
\section{Weight-sharing Neural Architecture Search}

Weight-sharing in neural architecture search is now prominent because of its efficiency \cite{brock2017smash,pham2018efficient,bender2018understanding,guo2019single}. They can roughly be divided into two categories.

\textbf{One-stage approaches.}
Specifically, ENAS \cite{pham2018efficient} adopts a reinforced approach to train a controller to sample subnetworks from the supernet. `Good' subnetworks yield high rewards so that the final policy of the controller is able to find competitive ones. Notice the controller and subnetworks are trained alternatively. DARTS \cite{liu2018darts} and many of its variants \cite{wu2018fbnet,xie2018snas,dong2019searching,chen2019progressive} are also a nested optimization based on weight-sharing but in a differentiable way. Besides, DARTS creates an exclusive competition by selecting only one operation on an edge, opposed to ENAS where more operations can be activated at the same time.

\textbf{Two-stage approaches.}
There are some other weight-sharing methods who use the trained supernet as an evaluator \cite{brock2017smash,bender2018understanding,guo2019single}. We need to make a distinction here as DARTS is a one-stage approach. The supernet of DARTS is meant to learn towards a single solution, where other paths are less valued (weighted). Instead like in \cite{guo2019single}, all paths are uniformly sampled, so to give an equal importance on selected paths. As the supernet is used for different purposes, two-stage approaches should be singled out for discussion in this paper. 


\section{Search Spaces}\label{sec:ss-supp}

There are two search spaces extensively adopted in recent NAS approaches. The one DARTS proposed, we denote as $S_1$, is cell-based with flexible inner connections \cite{liu2018darts,chen2019progressive}, and the other $S_2$ is at the block level of the entire network \cite{tan2018mnasnet,cai2018proxylessnas,wu2018fbnet}. We use the term \emph{proxy} and \emph{proxyless}  to differentiate whether it directly represents the backbone architecture. Our experiments cover both categories, if otherwise explicitly written, the first is searched on CIFAR-10 and the second on ImageNet. 

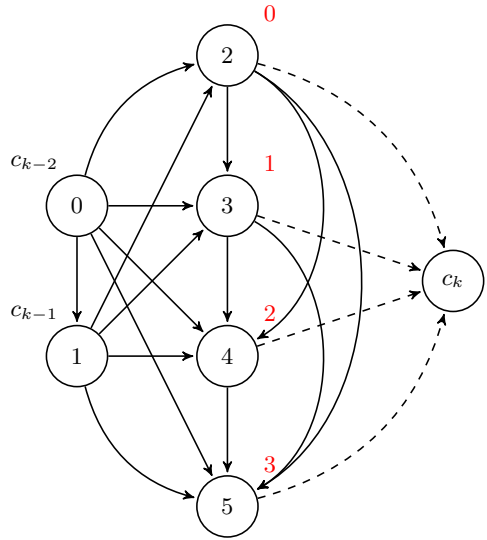
\begin{figure}[ht]
	\centering
	\begin{tikzpicture}[->,>=stealth',shorten >=1pt,auto,node distance=2cm,
	semithick]
	\node[state] (a)  {0};
	\node[state] (b) [below of=a] {1};
	\node[state] (d) [right of=a] {3};
	\node[state] (c) [above of=d] {2};
	\node[state] (e) [below of=d] {4};
	\node[state] (f) [below of=e] {5};
	\node[state] (g) [right of=e, node distance=3cm, yshift=1cm] {$c_k$};
	\path (a) edge  (b)
	(a) edge [bend left] (c)
	(a) edge (d)
	(a) edge (e)
	(a) edge  (f);
	\path (b) edge  (c)
	(b) edge (d)
	(b) edge (e)
	(b) edge[bend right] (f);
	\path (c) edge (d)
	(c) edge [bend left=60] (e)
	(c) edge [bend left=60] (f);
	\path (d) edge (e)
	(d) edge [bend left=60] (f); 
	\path(e) edge (f);
	\path[dashed] (c) edge[bend left] (g)
	(d) edge (g)
	(e) edge (g)
	(f) edge[bend right] (g);
	\node (A) [above right of=c, node distance=.8cm, text=red] {0};
	\node (B) [above right of=d, node distance=.8cm, text=red] {1};
	\node (C) [above right of=e, node distance=.8cm, text=red] {2};
	\node (D) [above right of=f, node distance=.8cm, text=red] {3};
	\node (E) [above left of=a, node distance=.8cm] {$c_{k-2}$};
	\node (F) [above left of=b, node distance=.8cm] {$c_{k-1}$};
	\end{tikzpicture}
	\caption{The DARTS search space at cell-level. Red labels indicate intermediate nodes. The outputs of all intermediate nodes are concatenated to form the output $c_k$}
	\label{fig:darts-ss}
\end{figure}

\textbf{Search Space $S_1$.} Our first search space $S_1$ (show in Fig.~\ref{fig:darts-ss}) follows DARTS \cite{liu2018darts} with an exception of excluding the \emph{zero operation}, as done in \cite{zela2020understanding}. Namely, $S_1$ works in the previously mentioned DAG of $N = 7$ nodes, first two nodes in cell $c_{k-1}$ are the outputs of last two cells $c_{k-1}$ and $c_{k-2}$, four intermediate nodes with each has incoming edges from the former nodes. The output node of DAG is a concatenation of all intermediate nodes. Each edge contains 7 candidate operations:

\begin{itemize}
	\item max\_pool\_3x3, avg\_pool\_3x3, skip\_connect,
	\item sep\_conv\_3x3, sep\_conv\_5x5,
	\item dil\_conv\_3x3, dil\_conv\_5x5.
\end{itemize}

\textbf{Search Space $S_2$.} The second search space $S_2$ is similar to that of ProxylessNAS  \cite{cai2018proxylessnas} which uses MobileNetV2 \cite{sandler2018mobilenetv2} as its backbone. We make several essential modifications. Specifically, there are $L=19$ layers and each contains $N=7$ following choices,

\begin{itemize}
	\item Inverted bottlenecks with an expansion rate $x$ in (3,6), a kernel size $y$ in (3,5,7), later referred to as E$x$K$y$, 
	\item skip connection.
\end{itemize}

\section{Experiment Details}

The list of all experiments we perform for this paper are summarized in Table~\ref{tab:all-experiments}. 

To summarize, we run DARTS \cite{liu2018darts} in $S_1$ and $S_2$, confirming the aggregation of skip connections in both search spaces. We also show the large discretization gap in $S_2$.

\begin{figure}[ht]
	\centering
	\includegraphics[width=0.4\textwidth,scale=0.6]{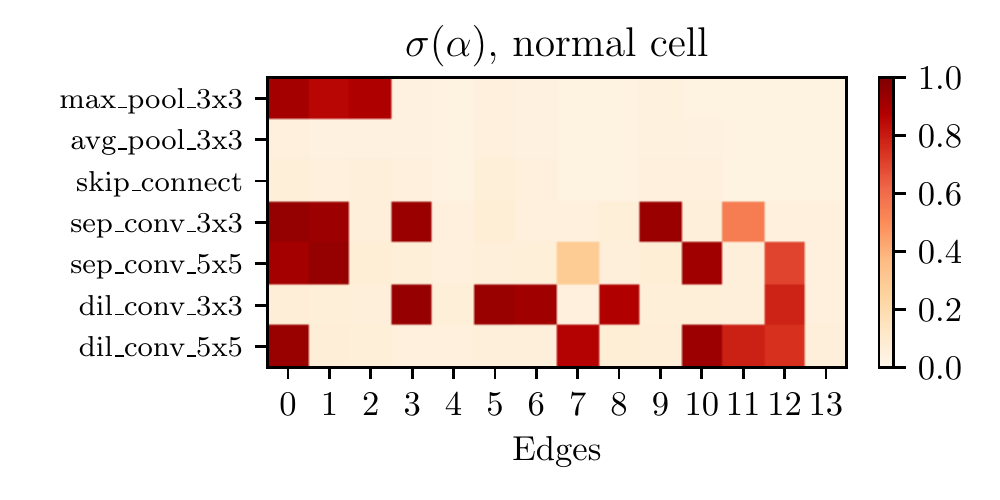}
	\includegraphics[width=0.4\textwidth,scale=0.6]{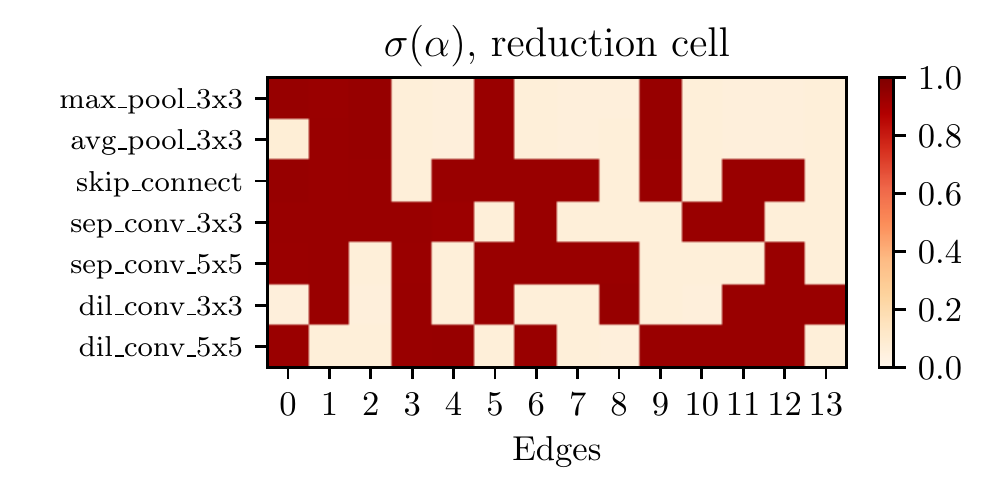}
	\caption{Heatmap of $\sigma(\alpha)$ in the normal cell and the reduction cell at the last epoch when searching with Fair DARTS on CIFAR-10 (in $S_1$). As a result of the sigmoid feature and the auxiliary loss $L_{0-1}$, the values are mainly around 0 and 1}
	\label{fig:sigmoid-fair-darts-cifar}
\end{figure}

\begin{figure}[ht]
	\centering
	\includegraphics[scale=0.6]{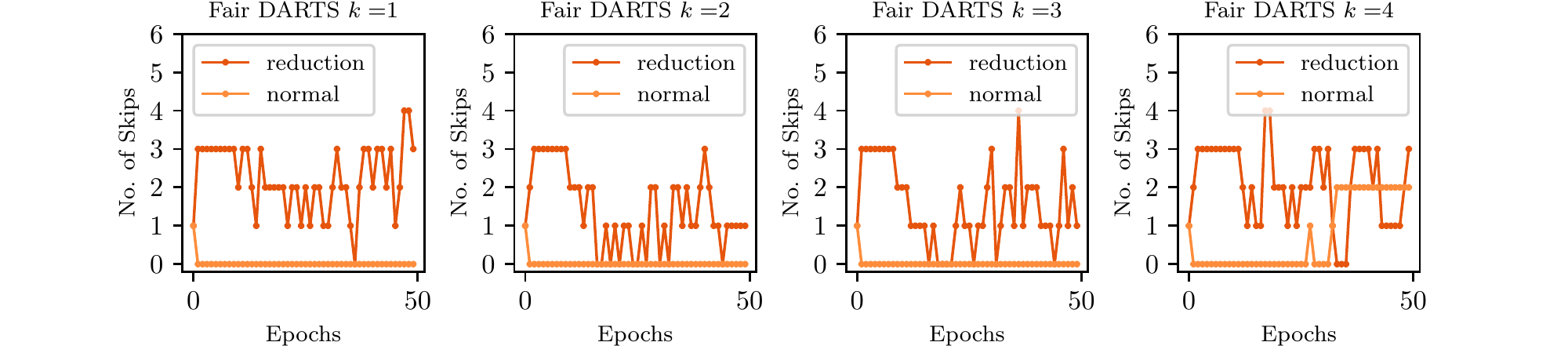}
	\caption{The number of dominant skip connections when searching with Fair DARTS on CIFAR-10. Here we select top-2 operations as dominants as in DARTS (our proposed way selects dominants according to $\sigma_{threshold}$), our method can still escape from too many skip connections. Note that we choose the first 4 experiments to show}
	\label{fig:num-skip-cifar-fair-darts}
\end{figure}

In comparison, we run Fair DARTS in $S_1$ and $S_2$ to show their differences. First, due to our collaborative approach, we can allow a substantial number of skip connections in $S_1$, see Fig.~\ref{fig:num-skip-cifar-fair-darts} (supplementary). Second, Fig.~\ref{fig:sigmoid-fair-darts-cifar} (supplementary) exhibit the final heatmap of architectural weights $\sigma$ where skip connections coexist with other operations, meantime, the values of $\sigma$  are also close to 0 or 1, which can minimize the discretization gap.

\subsection{Search on CIFAR-10 in Search Space $S_1$}
We initialize all architectural weights to 0\footnote{Its sigmoid output is 0.5 (fair starting point).} and set $w_{0-1}=10$ . We comply with the same hyperparameters with some exceptions: a batch size 128, learning rate 0.005, and Adam optimizer with a decay of 3e-3 and beta (0.9, 0.999).  Moreover, we comply with the same strategy of grouping training and validation data as \cite{liu2018darts}. The edge correspondence in search space $S_1$ is given in Table~\ref{tab:darts-edge-s1}. We also use the first-order optimization to save time. 

\begin{figure}[ht]
	\centering
	\includegraphics[scale=0.38]{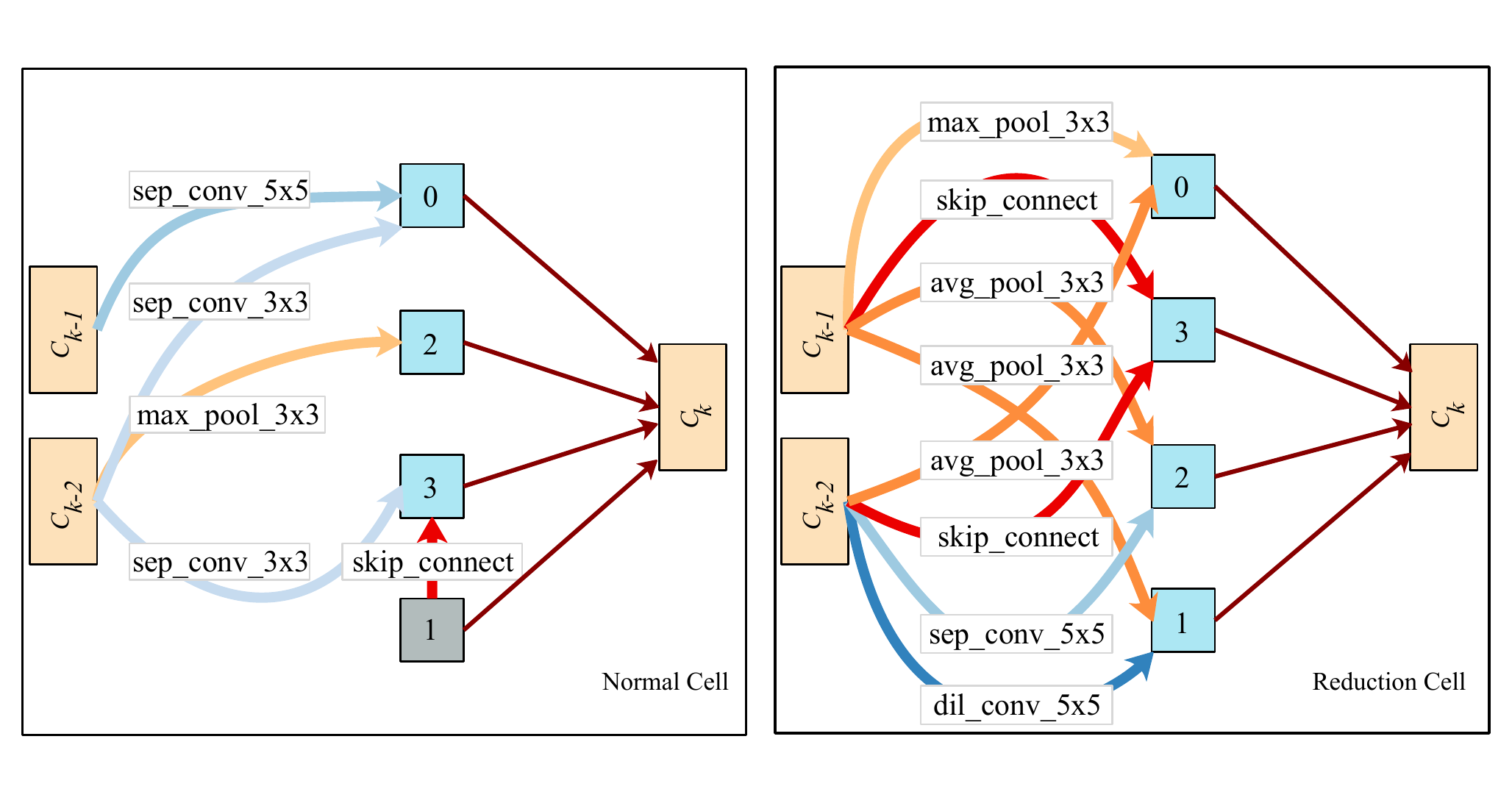}
	\caption{FairDARTS-a cells searched on CIFAR-10 in $S_1$}
	\label{fig:normal-reduce-architecture}
\end{figure}

\begin{figure}[ht]
	\centering
	\includegraphics[scale=0.38]{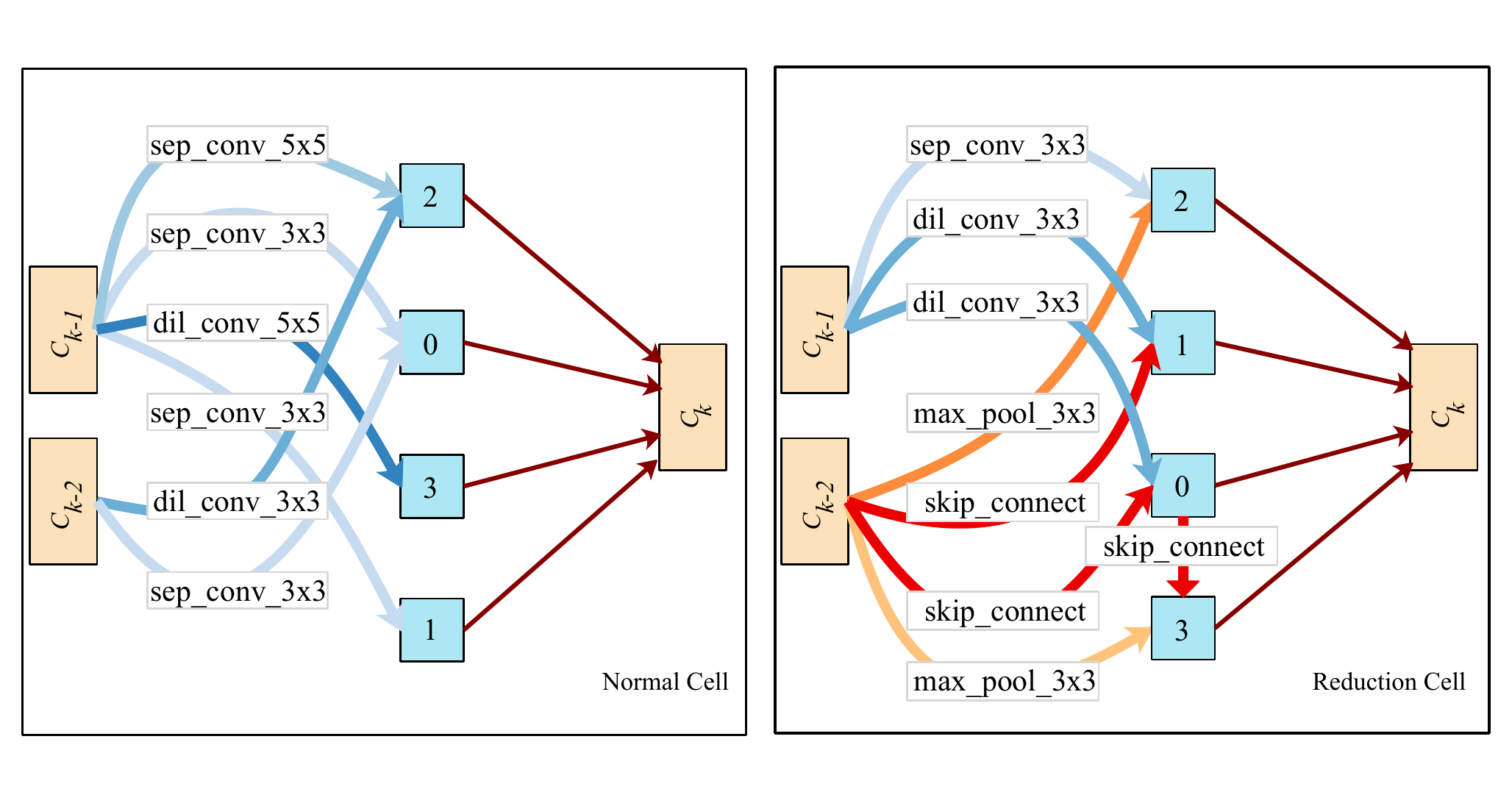}
	\caption{FairDARTS-b cells searched on CIFAR-10 in $S_1$}
	\label{fig:fairdarts-b-normal-reduce-architecture}
\end{figure}

We illustrate some of the searched cells in Fig.~\ref{fig:normal-reduce-architecture} and Fig.~\ref{fig:fairdarts-b-normal-reduce-architecture}. The normal cell of FairDARTS-a is rather simple and only 3 nodes are activated, which is previously rarely seen. In the reduction cell, more edges are activated to compensate for the information loss due to down-sampling. As reduction cells are of small proportion, FairDARTS-a thus retains to be lightweight. Other searching results are listed in Table~\ref{tab:fairdarts_models}.

\subsection{Search on ImageNet in Search Space $S_2$}

We use the SGD optimizer with an initial learning rate 0.02 (cosine decayed by epochs) for the network weights and Adam optimizer with 5e-4 for architectural weights.   Besides, we set $\sigma_{threshold}$ = 0.75 and $w_{0-1}$ = 1.0.


There is a minor issue need to concern. The innate \emph{skip connections} are removed from inverted bottlenecks since skip connections have already been made a parallel choice in our search space.  

\begin{table*}
	\begin{center}
		\caption{All experiments conducted for this paper. By default, $S_1$ is for CIFAR-10, $S_2$ for ImageNet. Note $\mathcal{L}_{val}$ refers to $\mathcal{L}_{val} (w^*(\alpha), \alpha)$. $^\dagger$: Main text. $^\ddagger$: In the supplementary. $^\star:\mathcal{N}(0,1)$, run $k=4$ times, $^*:$$w_{0-1} \in range(16)$}
		\label{tab:all-experiments}
		\setlength{\tabcolsep}{2mm}{
			\resizebox{\textwidth}{20mm}{
				\begin{tabular}{llp{3cm}*{5}{c}}
					\hline
					Idx & Method & Search Space & Loss & Optimization & Fig$^\dagger$  & Fig$^\ddagger$ & Table$^\ddagger$ \\
					\hline
					1 & DARTS$^\star$ & $S_1$ & $\mathcal{L}_{val}$ & Bi-level & \ref{fig:num-skip-cifar},\ref{fig:darts-bar-skip-dominant-edge},\ref{fig:alpha-heatmap-darts-cifar-imagenet} & \ref{fig:darts-softmax-all-s1},\ref{fig:darts-3-reduce-skip-bar-per-edge} &\\
					2 & DARTS & $S_2$ & $\mathcal{L}_{val}$ & Bi-level &\ref{fig:num-skip-imagenet},\ref{fig:alpha-heatmap-darts-cifar-imagenet} &&\\
					3 & Fair DARTS & $S_1$ & $\mathcal{L}_{val}$ & Bi-level & \ref{fig:histo-loss-and-weight-dominant}&&  \\
					4 & Fair DARTS$^*$ & $S_1$ & $\mathcal{L}_{val} + w_{0-1}L_{0-1}$ & Bi-level & \ref{fig:histo-loss-and-weight-dominant} & & \\
					5 & Fair DARTS$^\star$ & $S_1$ & $\mathcal{L}_{val} + 10 * L_{0-1}$ & Bi-level & \ref{fig:histo-loss-and-weight-dominant} & \ref{fig:single-bilevel-sigmoid-comparison},\ref{fig:sigmoid-fair-darts-cifar},\ref{fig:normal-reduce-architecture},\ref{fig:fairdarts-b-normal-reduce-architecture},\ref{fig:bilevel-alpha},\ref{fig:darts-knight-bar-skip-dominant-edge-all-reduce} & \ref{tab:fairdarts_models},\ref{fig:num-skip-cifar-fair-darts}  \\
					6 & Fair DARTS & $S_1$ & $\mathcal{L}_{val} + 10 * L'_{0-1}$ & Bi-level &  & \ref{fig:l1-l2-sigmoid-comparison},\ref{fig:l1-alpha}  &\\
					7 & Fair DARTS & $S_1$ & $\mathcal{L}_{val} + 10*L_{0-1}$ & Single-level & & \ref{fig:single-bilevel-sigmoid-comparison},\ref{fig:single-alpha}&\\
					8 & Fair DARTS & $S_2$ & $\mathcal{L}_{val} + 10*L_{0-1}$ & Bi-level & \ref{fig:num-skip-imagenet},\ref{fig:fair-darts-arch} && \\
					9 & DARTS & $S_1 \setminus \{\text{skips}\}$ &  $\mathcal{L}_{val}$ & Bi-level &\ref{fig:darts-bar-no-skip-dominant-edge} & \ref{fig:darts-no-skip-bar-dominant-edge-all-reduce}& \\
					10 & Random & $S_1 \cap \{\text{priors}\}$ & n/a & n/a & & & \ref{tab:random-m2noflops} \\
					11 & Random & $S_1 \cap \{\text{priors}\} \cap \{\text{FLOPS} \geq 500M\}$ & n/a & n/a & && \ref{tab:random-m2500flops} \\
					12 & Noise$^\star$ & $S_1 $ &  $\mathcal{L}_{val}$ & Bi-level & && \ref{tab:noisy-darts-lr-decay} \\
					13 & DARTS & $S_1$ & $\mathcal{L}_{val} + 10*L_{0-1}$ & Bi-level & & \ref{fig:darts-l2-skip-no-acc},\ref{fig:darts-l2-ops-alpha}& \\
					\hline
		\end{tabular}}}
	\end{center}
\end{table*}

\begin{table}
	\begin{center}
		\caption{Edge correspondence in $S_1$. Edge $i$ is represented as a pair $(j, k)$ where $j$ is an intermediate node, and $k$ is the incoming node. Note $j$ is numbered only on intermediate nodes. And $k$ counts previous cell outputs as well. See Fig.~\ref{fig:darts-ss} for correspondence}
		\label{tab:darts-edge-s1}
		\begin{footnotesize}
			\begin{tabular}{cl}
				\hline
				Edge & $(j, k)$  \\
				\hline
				0 & (\textcolor{red}{0}, 0)\\
				1 & (\textcolor{red}{0}, 1)\\
				2 & (\textcolor{red}{1}, 0)\\
				3 & (\textcolor{red}{1}, 1)\\
				4 & (\textcolor{red}{1}, 2)\\
				5 & (\textcolor{red}{2}, 0)\\
				6 & (\textcolor{red}{2}, 1)\\
				7 & (\textcolor{red}{2}, 2)\\
				8 & (\textcolor{red}{2}, 3)\\
				9 & (\textcolor{red}{3}, 0)\\
				10 & (\textcolor{red}{3}, 1)\\
				11 & (\textcolor{red}{3}, 2)\\
				12 & (\textcolor{red}{3}, 3)\\
				13 & (\textcolor{red}{3}, 4)\\
				\hline
			\end{tabular}
		\end{footnotesize}
	\end{center}
	
\end{table}

\begin{table}
	\begin{center}
		\caption{Dropout strategies of DARTS variants}
		\label{tab:dropout}
		\begin{footnotesize}
			\begin{tabular}{lcl}
				\hline
				Mothods & Skip Connection  & Other Ops \\
				\hline
				DARTS \cite{liu2018darts} & No Drop & No Drop  \\
				PDARTS \cite{chen2019progressive} &Drop & No Drop \\
				RobustDARTS \cite{zela2020understanding} & Drop & Drop \\
				\hline
			\end{tabular}
		\end{footnotesize}
	\end{center}
\end{table}

\subsection{Zero-one Loss Comparison}
\begin{figure}[ht]
	\centering
	\includegraphics[scale=0.6]{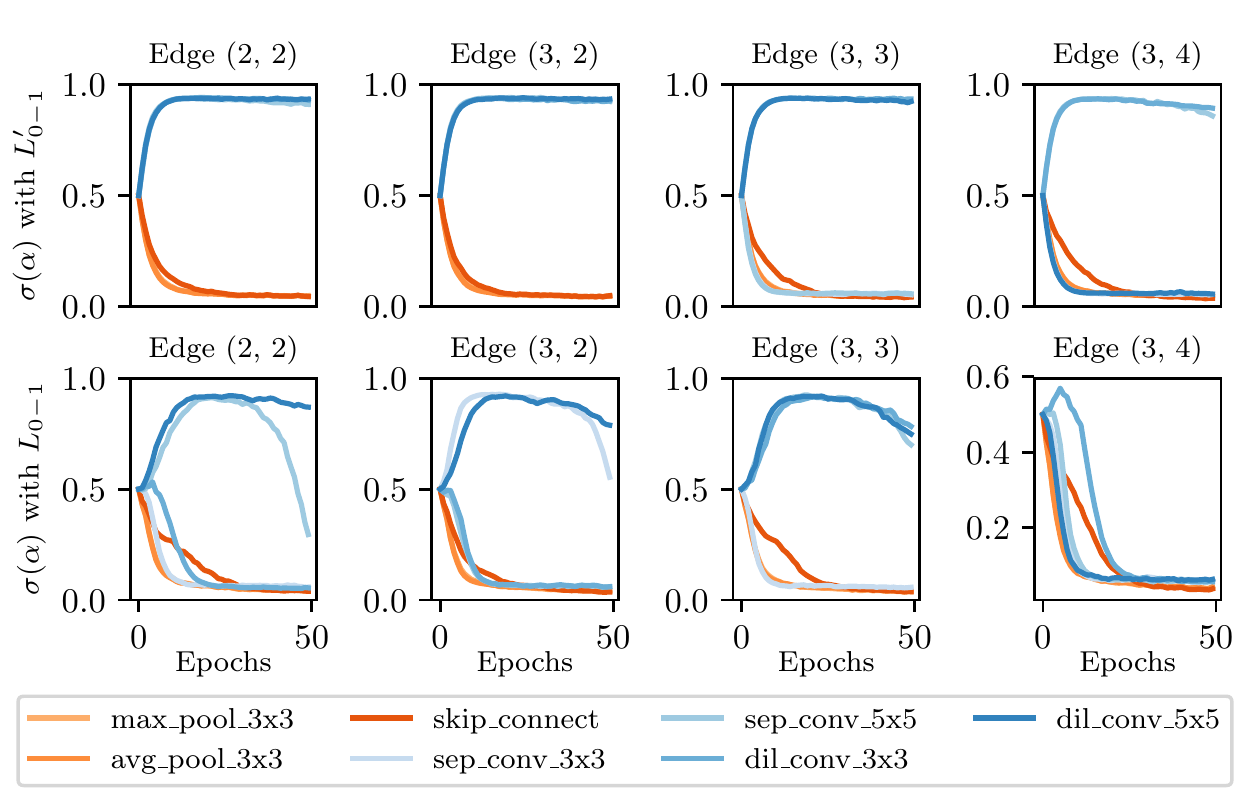}
	\caption{Comparison of sigmoid evolution when running Fair DARTS with $L'_{0-1}$ (top) and $L_{0-1}$ (bottom) in $S_1$. With the proposed loss at the bottom, sigmoid values manage to step out of local optima }
	\label{fig:l1-l2-sigmoid-comparison}
\end{figure}

Fig.~\ref{fig:l1-l2-sigmoid-comparison} compares results on two different loss designs. With the proposed loss function $L_{0-1}$ so that Fair DARTS is less subjective to initialization. The sigmoid values reach to their optima more slowly than that of $L'_{0-1}$. It also gives a fair opportunity near 0.5,  the 3$\times$3 dilation convolution on Edge (3,4) first increases and then decreases, which matches the second criteria of loss design.

\subsection{Single-level vs. Bi-level Optimization}
\begin{figure}[ht]
	\centering
	\includegraphics[scale=0.6]{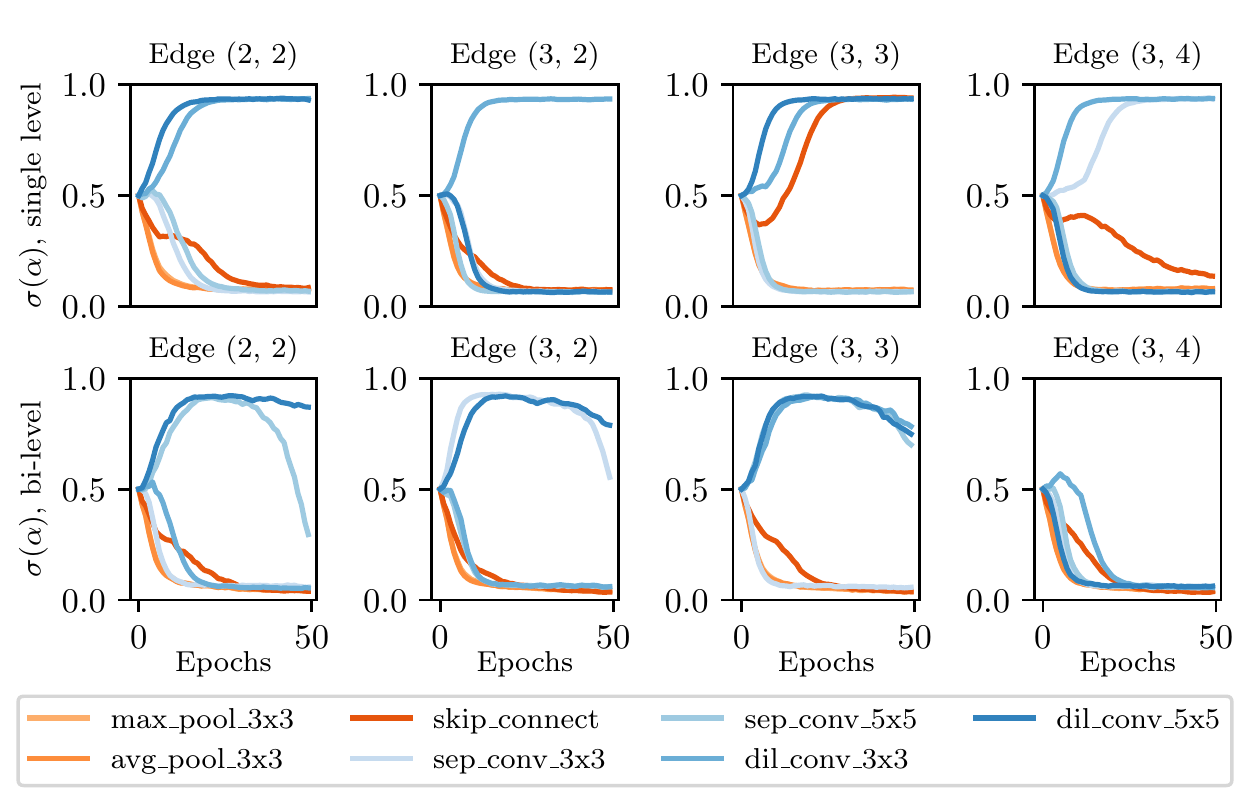}
	\caption{Comparison of sigmoid evolution when running Fair DARTS with single and bi-level optimization. Bi-level has better robustness to local minimum}
	\label{fig:single-bilevel-sigmoid-comparison}
\end{figure}

The $5\times5$ separable convolution on edge (2, 2) under bi-level setting weighs higher at the early stage but much decreased in the end, which can be viewed as robustness to a local optimum. See Fig.~\ref{fig:single-bilevel-sigmoid-comparison}.

\subsection{Results on COCO Object Detection}

We use our models as  drop-in replacements of the backbone of RetinaNet \cite{lin2017focal}. Here we only consider comparable mobile backbones. Particularly, we use the MMDetection \cite{chen2019mmdetection} toolbox since it provides good implementations for various objective methods.  All the models are trained and evaluated on MS COCO dataset (train2017 and val2017 respectively)  for 12 epochs with a batch size of 16. The initial learning rate is 0.01 and divided by 10 at epochs 8 and 11.  Table~\ref{table:fairdarts-coco-retina} shows that our model achieves the best average precision $31.9\%$. 

\setlength{\tabcolsep}{4pt}
\begin{table}
	\begin{center}
		\caption{Object detection of various drop-in backbones. $^{\dagger}$: w/ SE and Swish}
		\label{table:fairdarts-coco-retina}
		\begin{tabular}{llllllllll}
			\hline\noalign{\smallskip}
			Backbones & $\times +$  & Params &Acc    & AP & AP$_{50}$ & AP$_{75}$ & AP$_S$ & AP$_M$ & AP$_L$ \\
			& (M) & (M) & (\%) &(\%) & (\%)& (\%)&(\%) &(\%) &(\%) 
			\\
			\hline\noalign{\smallskip}
			MobileNetV2 \cite{sandler2018mobilenetv2} & 300 & 3.4& 72.0 & 28.3 & 46.7 & 29.3 & 14.8 & 30.7 & 38.1\\
			SingPath NAS \cite{stamoulis2019single} & 365 & 4.3 & 75.0 & 30.7 & 49.8 & 32.2 & 15.4 &33.9 & 41.6\\
			MobileNetV3 \cite{howard2019searching} & 219 & 5.4 & 75.2& 29.9 & 49.3 & 30.8 & 14.9 & 33.3 & 41.1\\
			MnasNet-A2 \cite{tan2018mnasnet} & 340& 4.8 & 75.6 & 30.5 & 50.2 & 32.0 & 16.6 & 34.1 & 41.1\\
			MixNet-M \cite{tan2020mixconv} & 360 & 5.0 & 77.0 & 31.3& 51.7 & 32.4& 17.0 & 35.0 & 41.9   \\
			FairDARTSC  $^{\dagger}$ & 386 &5.3 & 77.2 & 31.9 & 51.9 & 33.0 & 17.4 & 35.3 & 43.0  \\
			\hline
		\end{tabular}
	\end{center}
\end{table}

\section{Figures}

\subsection{Evolution of Architectural Weights}
\begin{figure*}[ht]
	\centering
	\subfigure[Normal Cell]{
		\includegraphics[scale=0.45]{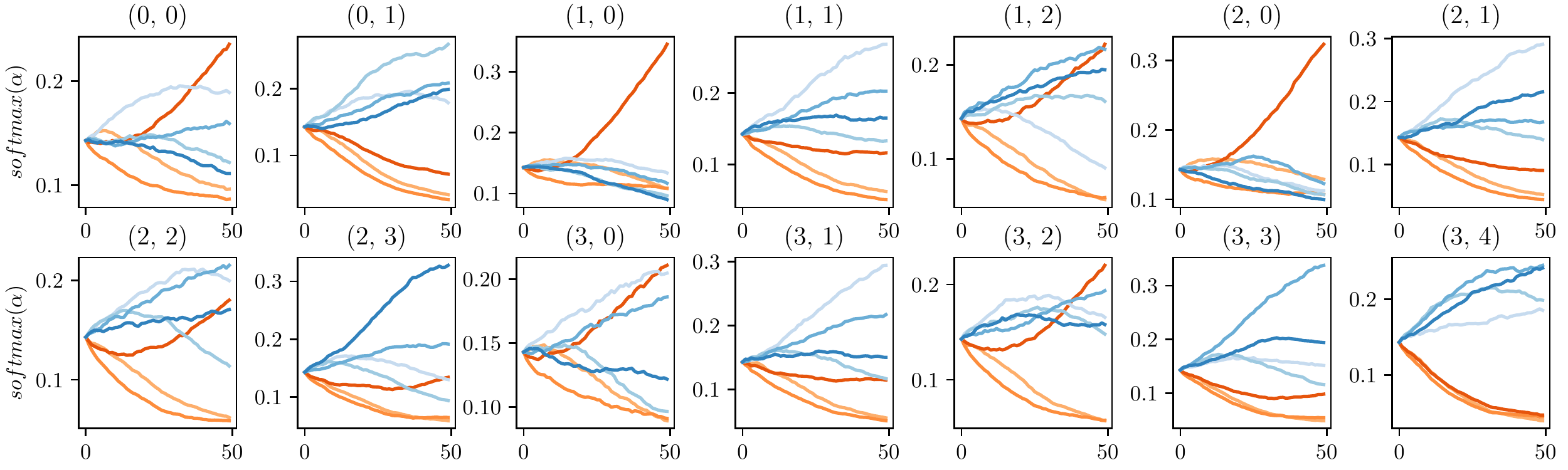}
	}
	\subfigure[Reduction Cell]{
		\includegraphics[scale=0.45]{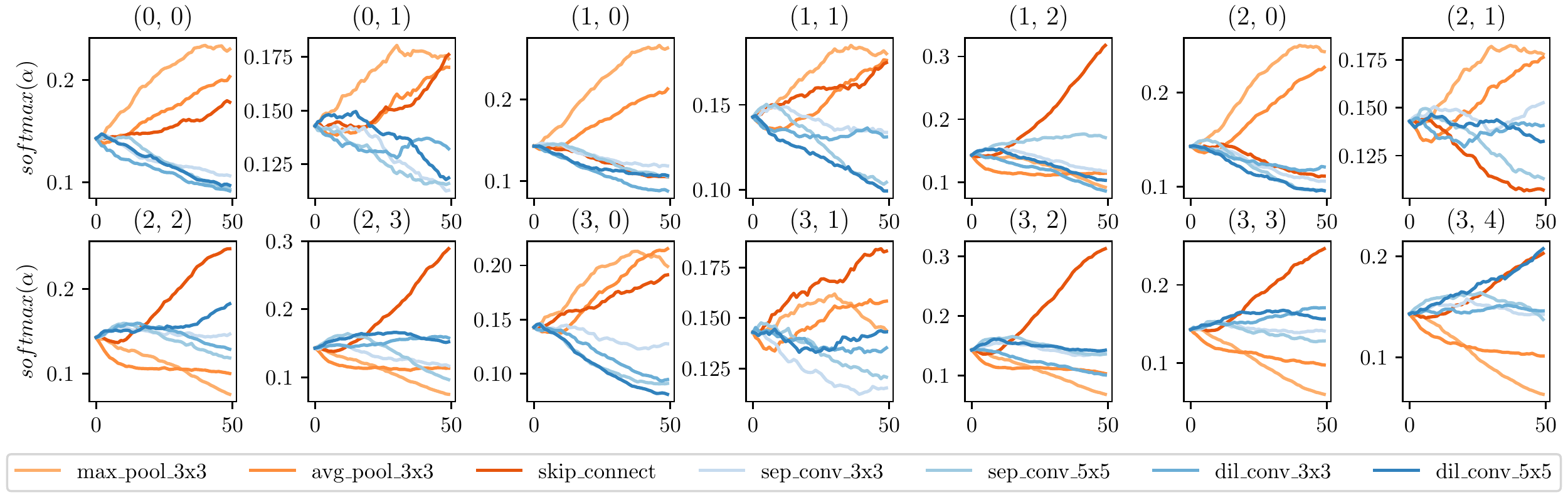}
	}
	\caption{The softmax evolution when running DARTS on CIFAR-10 in $S_1$ ($k=3$)}
	\label{fig:darts-softmax-all-s1}
\end{figure*}

\begin{figure*}[ht]
	\centering
	\subfigure[Normal Cell]{
		\includegraphics[scale=0.48]{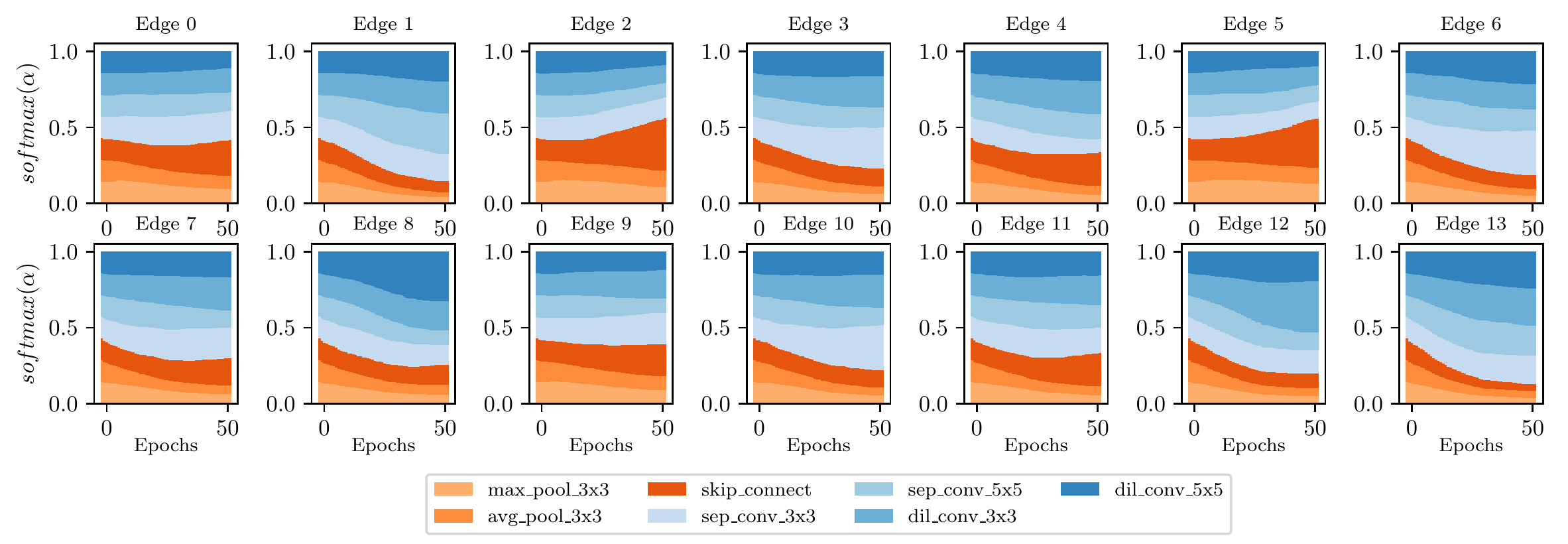}
	}
	\subfigure[Reduction Cell]{
		\includegraphics[scale=0.48]{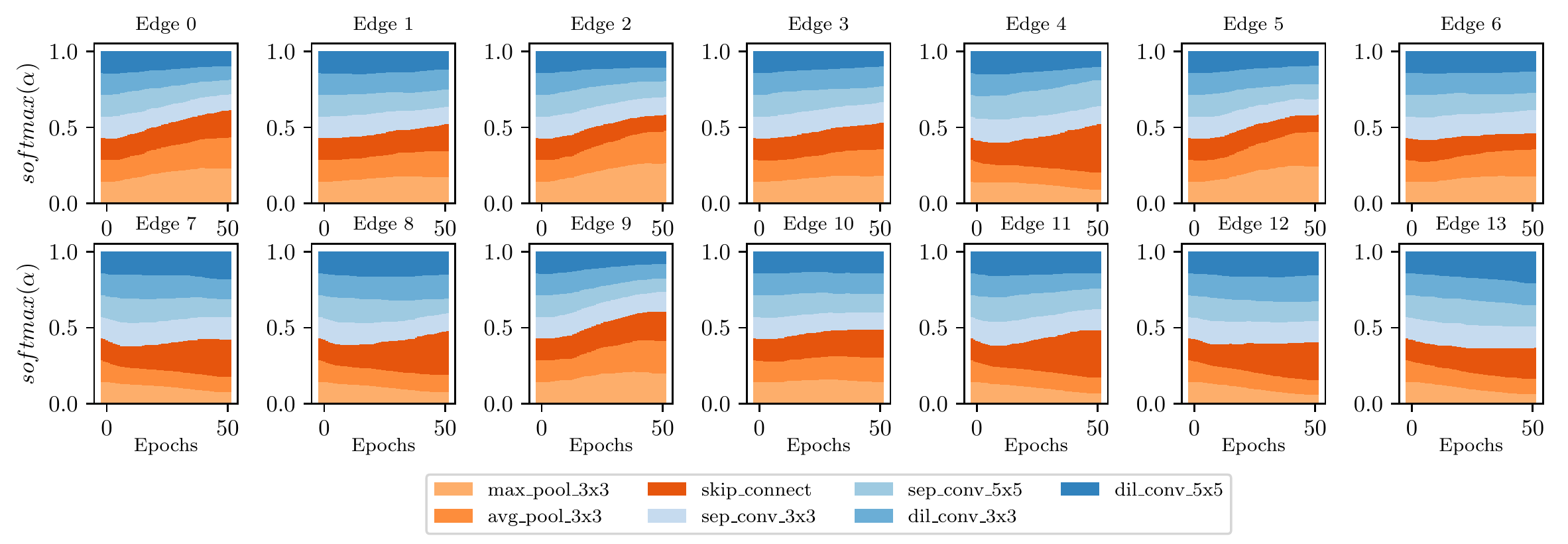}
	}
	\caption{The evolution of $softmax(\alpha)$ when running DARTS on CIFAR-10 in $S_1$. Skip connections on edge 0,2,4,5,11 in the normal cell and edge 4,7,8,11,12 in the reduction cell gradually suppress others caused by unfair advantage}
	\label{fig:darts-3-reduce-skip-bar-per-edge}
\end{figure*}

\begin{figure*}[ht]
	\centering
	\subfigure[Normal Cell]{
		\includegraphics[scale=0.48]{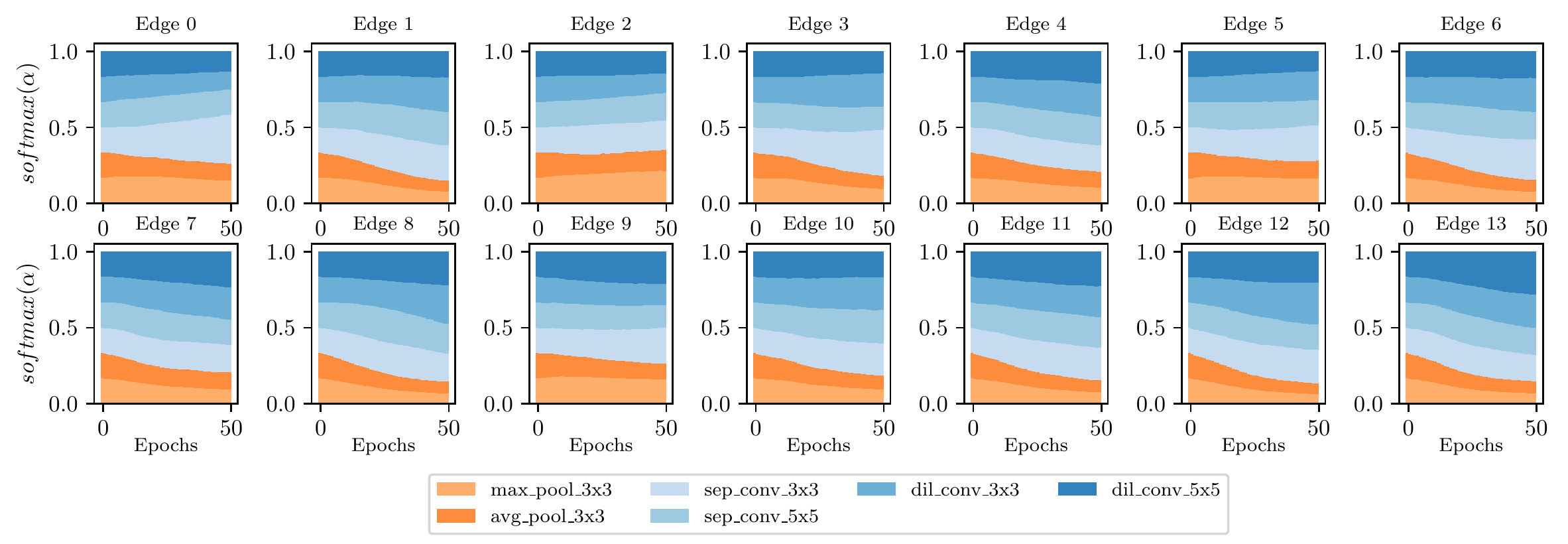}
	}
	\subfigure[Reduction Cell]{
		\includegraphics[scale=0.48]{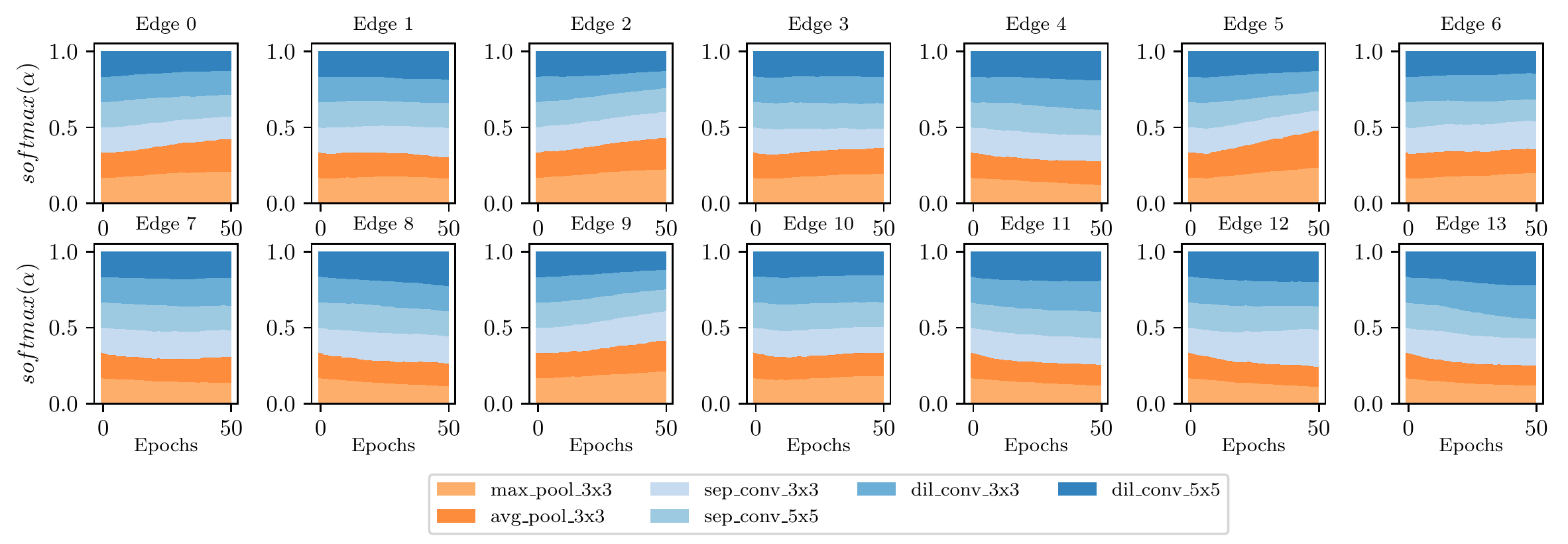}
	}
	\caption{The evolution of $\sigma(\alpha)$ when running DARTS on CIFAR-10 in $S_1$ without skip connections. With unfair advantages removed, all operations are encouraged to demonstrate each real strength}
	\label{fig:darts-no-skip-bar-dominant-edge-all-reduce}
\end{figure*}

Fig.~\ref{fig:darts-softmax-all-s1} and \ref{fig:darts-3-reduce-skip-bar-per-edge} gives the complete \emph{softmax} evolution when running DARTS on CIFAR-10 in $S_1$. Fig.~\ref{fig:darts-no-skip-bar-dominant-edge-all-reduce} is a similar case except the skip connection is removed. 

\begin{figure*}[ht]
	\centering
	\subfigure[Normal Cell]{
		\includegraphics[scale=0.45]{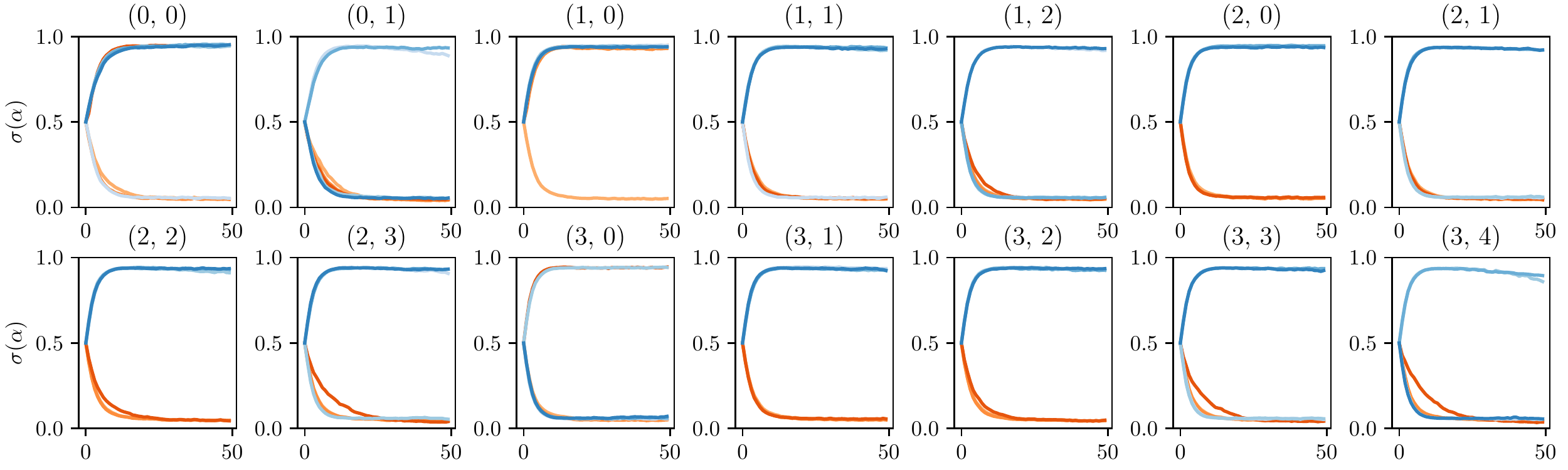}
	}
	\subfigure[Reduction Cell]{
		\includegraphics[scale=0.45]{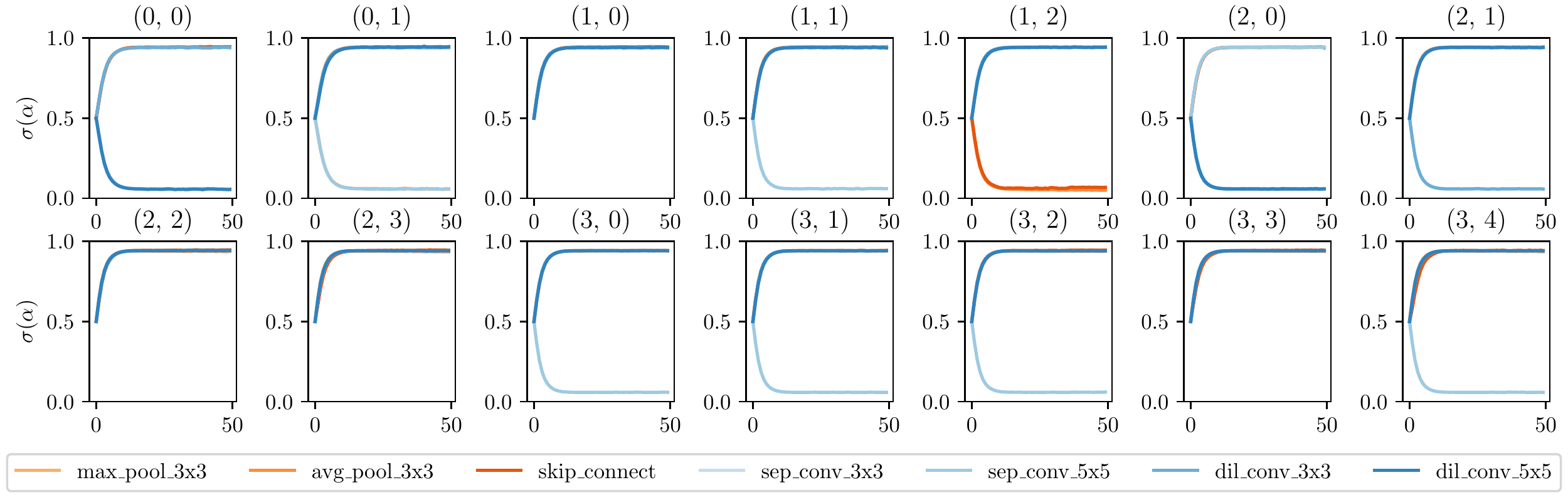}
	}
	\caption{The sigmoid evolution when running Fair DARTS with $L'_{0-1}$ loss in $S_1$}
	\label{fig:l1-alpha}
\end{figure*}

Fig.~\ref{fig:l1-alpha} gives the complete \emph{sigmoid} evolution when running Fair DARTS on CIFAR-10 in $S_1$ with $L'_{0-1}$.

\begin{figure*}[ht]
	\centering
	\subfigure[Normal Cell]{
		\includegraphics[scale=0.45]{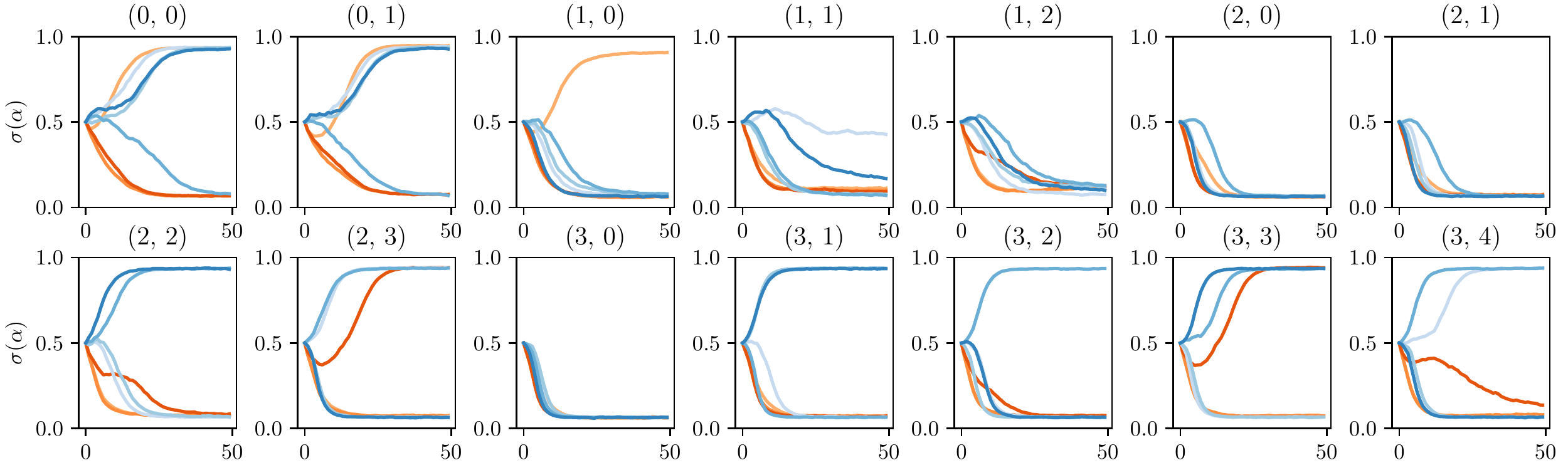}
	}
	\subfigure[Reduction Cell]{
		\includegraphics[scale=0.45]{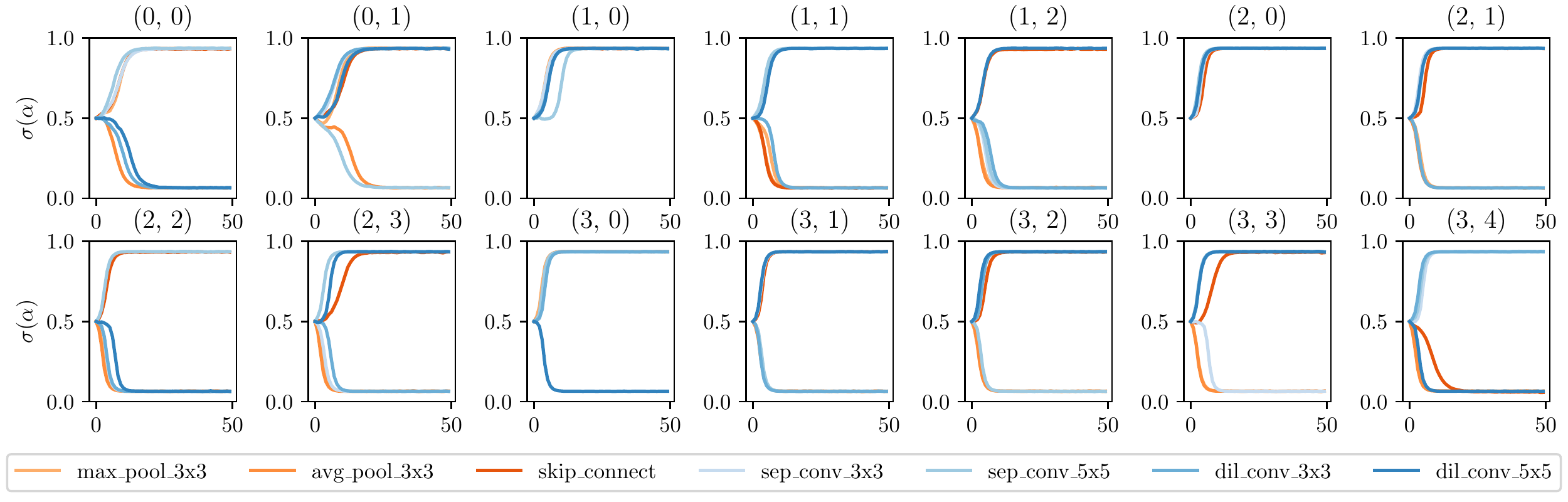}
	}
	\caption{The sigmoid evolution when running Fair DARTS with single-level optimization in $S_1$}
	\label{fig:single-alpha}
\end{figure*}

\begin{figure*}[ht]
	\centering
	\subfigure[Normal Cell]{
		\includegraphics[scale=0.45]{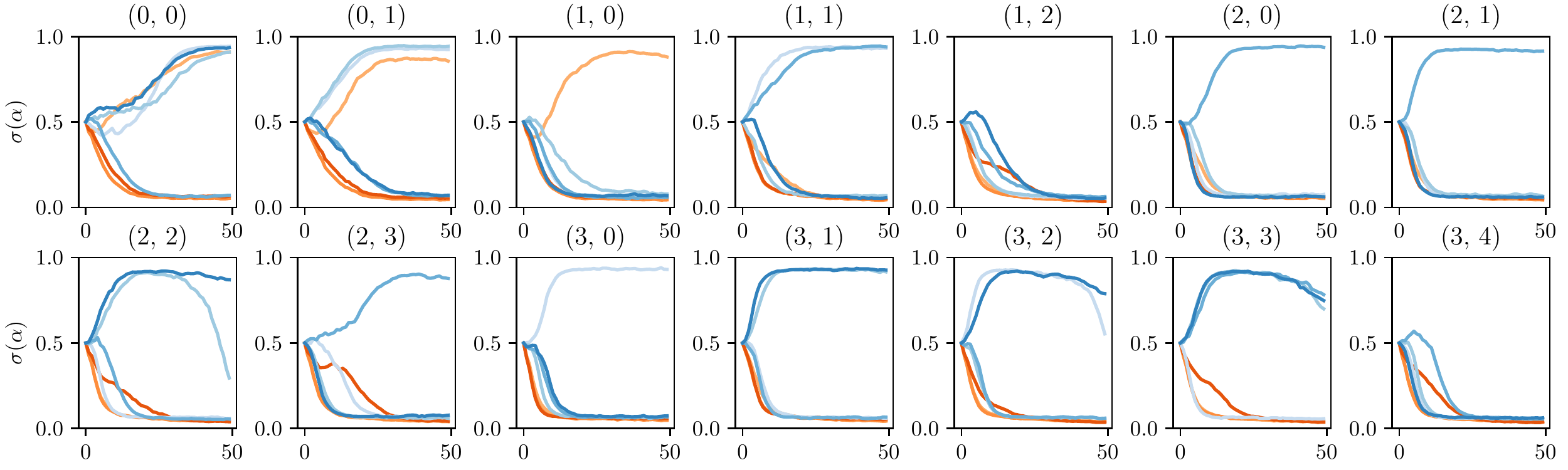}
	}
	\subfigure[Reduction Cell]{
		\includegraphics[scale=0.45]{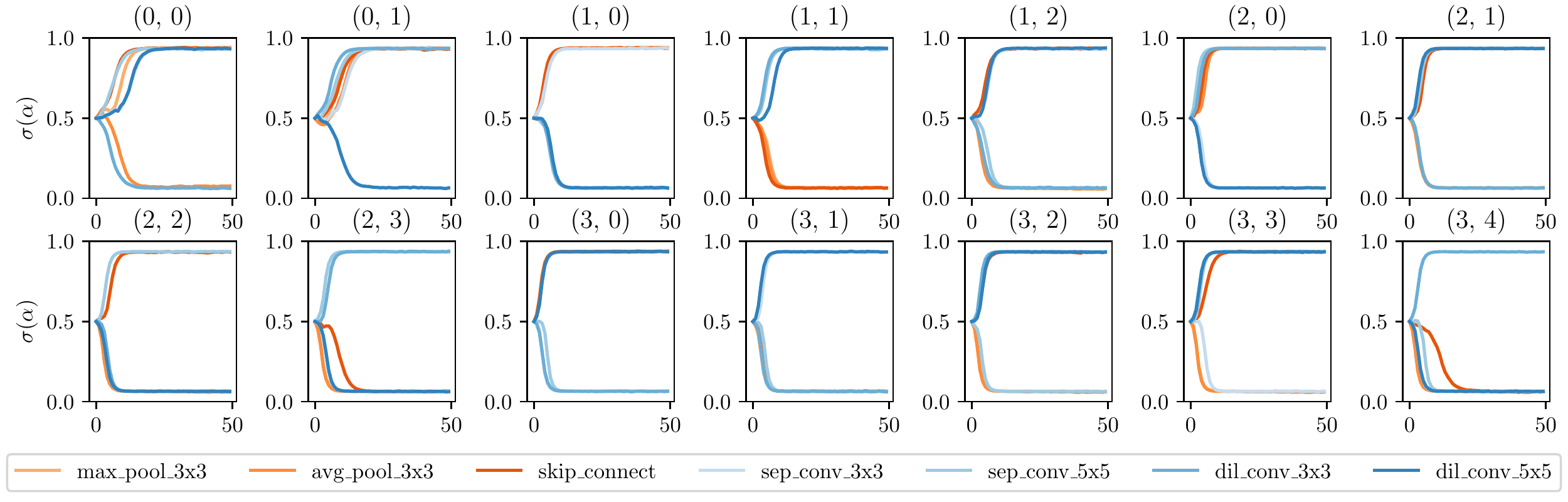}
	}
	\caption{The sigmoid evolution when running Fair DARTS with bi-level optimization in $S_1$}
	\label{fig:bilevel-alpha}
\end{figure*}

\begin{figure*}[ht]
	\centering
	\subfigure[Normal Cell]{
		\includegraphics[scale=0.48]{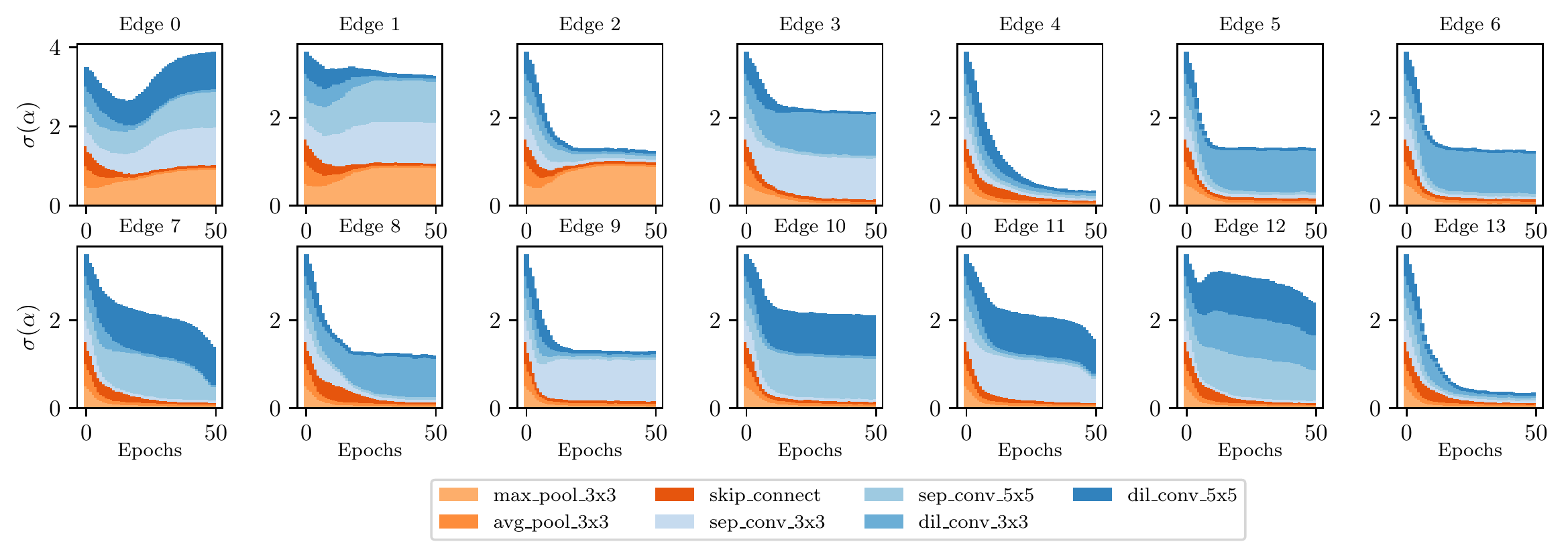}
	}
	\subfigure[Reduction Cell]{
		\includegraphics[scale=0.48]{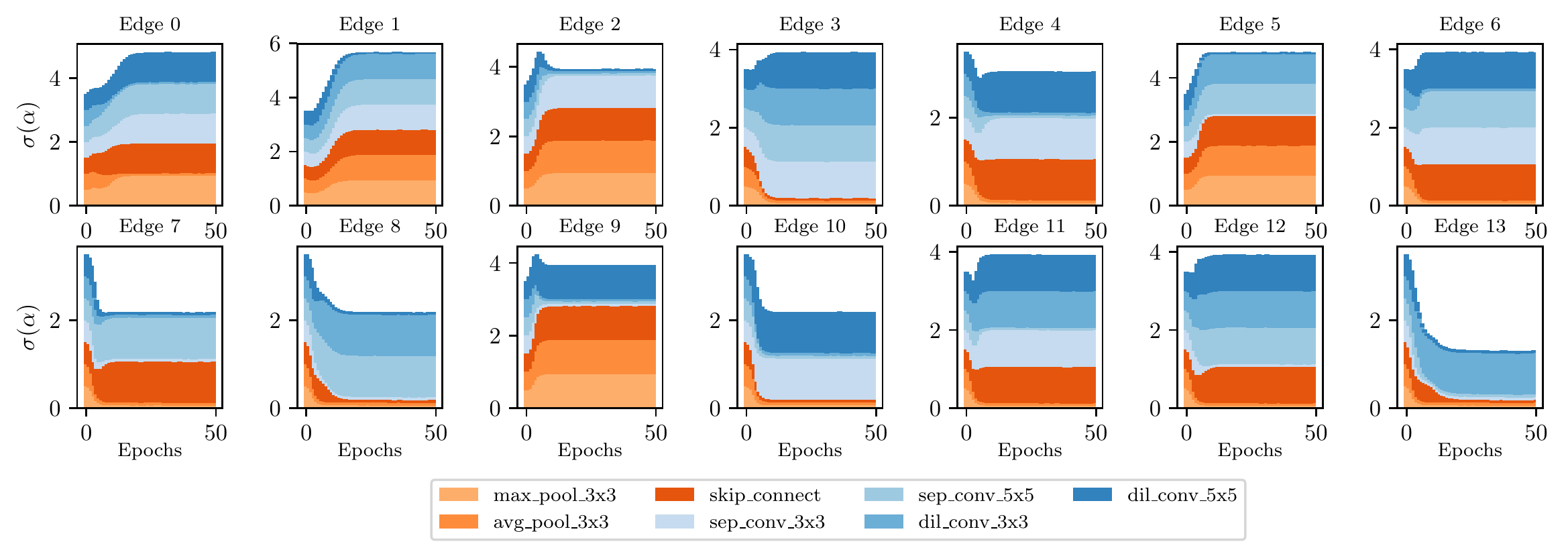}
	}
	\caption{The evolution of $\sigma(\alpha)$ when running Fair DARTS on CIFAR-10 in $S_1$. Skip connections enjoy an equal opportunity under collaborative competition. See edge 1,8,12 in normal cell and almost all edges in reduction cell}
	\label{fig:darts-knight-bar-skip-dominant-edge-all-reduce}
\end{figure*}

Fig.~\ref{fig:single-alpha}, \ref{fig:bilevel-alpha} gives the complete sigmoid evolution when running Fair DARTS on CIFAR-10 in $S_1$ with \emph{single-level} and \emph{bi-level} optimization respectively. Fig.~\ref{fig:darts-knight-bar-skip-dominant-edge-all-reduce} is a stacked-bar version of Fig.~\ref{fig:bilevel-alpha}.

\subsection{A Special Study: DARTS with $L_{0-1}$ }

\begin{figure}[ht]
	\centering
	\includegraphics[scale=0.6]{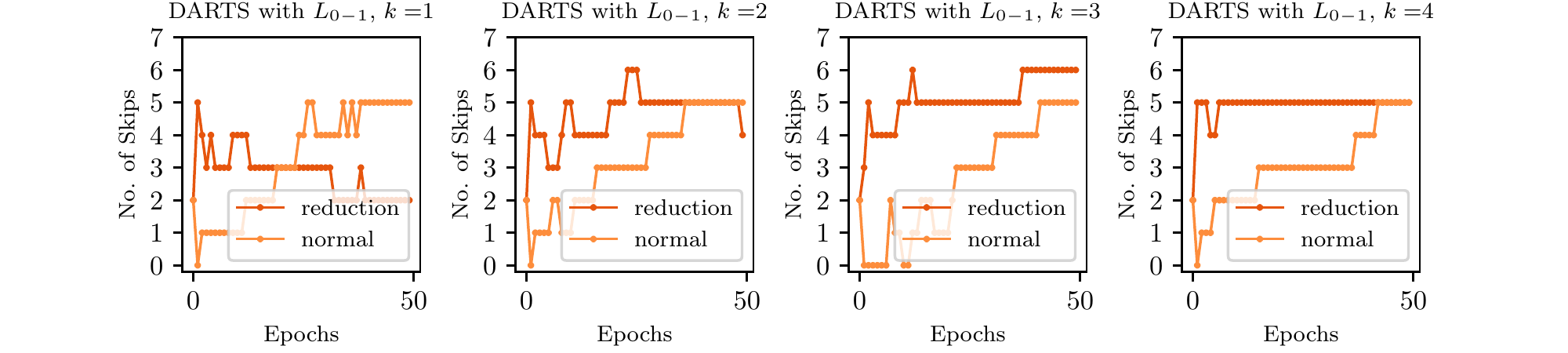}
	\caption{The number of dominant skip connections when searching with DARTS equipped with $L_{0-1}$ on CIFAR-10. Notice the aggregation of skip connections gets even worse}
	\label{fig:darts-l2-skip-no-acc}
\end{figure}

\begin{figure*}[ht]
	\centering
	\subfigure[Normal Cell]{
		\includegraphics[scale=0.48]{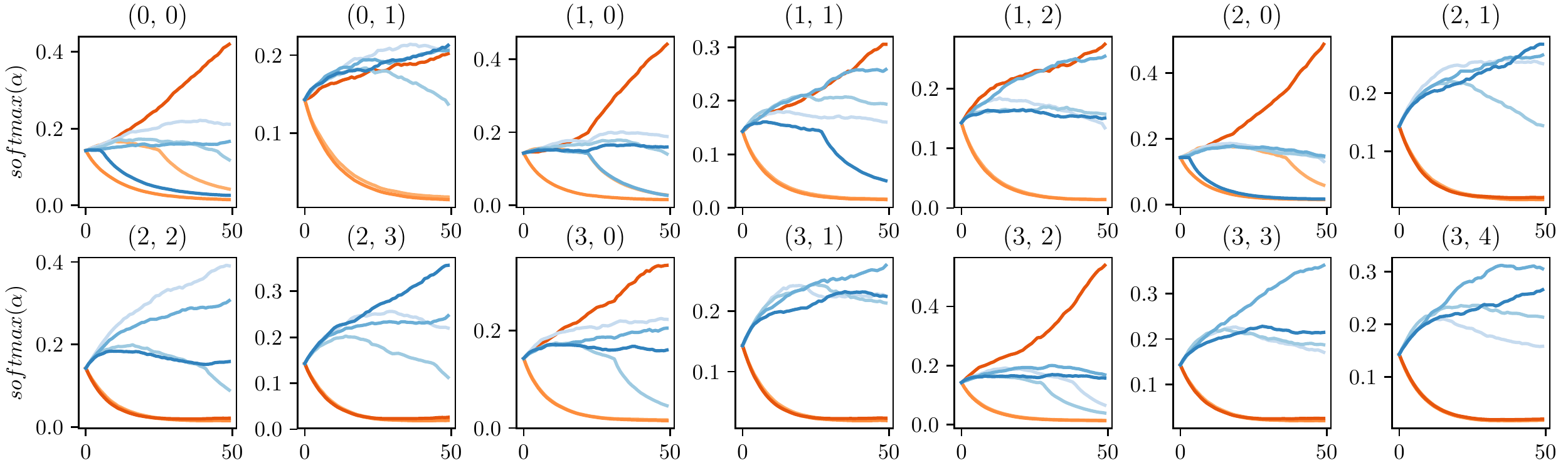}
	}
	\subfigure[Reduction Cell]{
		\includegraphics[scale=0.48]{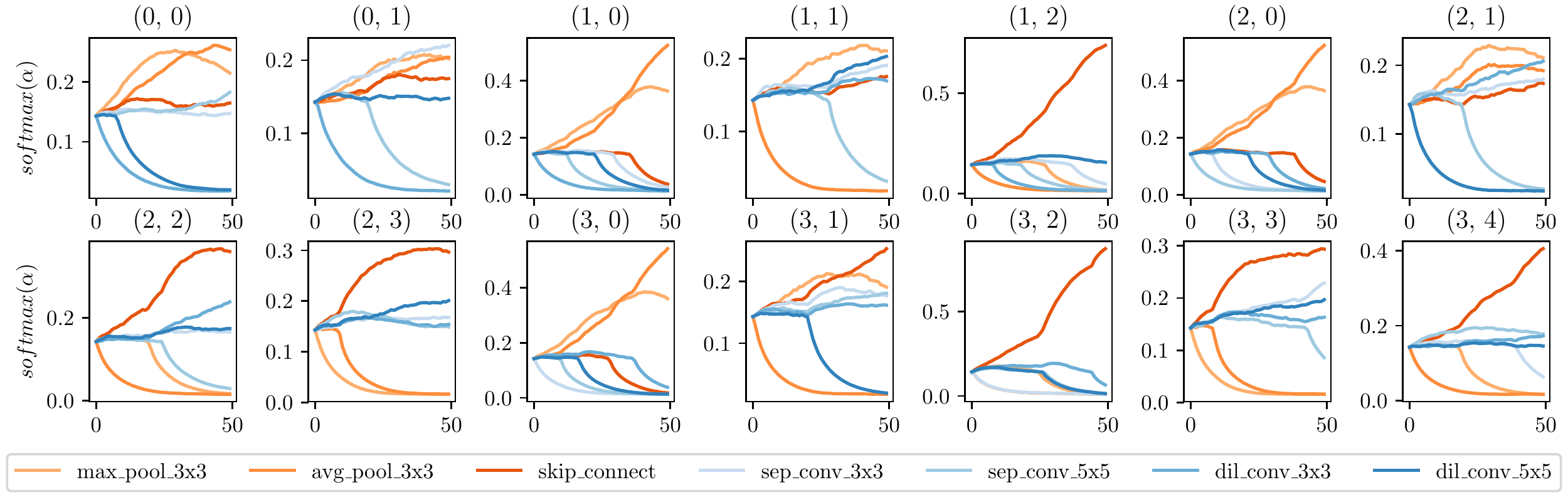}
	}
	\caption{The evolution of $softmax(\alpha)$ when running DARTS with the auxiliary loss $L_{0-1}$ enabled on CIFAR-10 in $S_1$}
	\label{fig:darts-l2-ops-alpha}
\end{figure*}

Although our auxiliary loss $L_{0-1}$ is particularly designed for Fair DARTS, it is interesting to see how DARTS behave under this new condition. Fig.~\ref{fig:darts-l2-skip-no-acc} and \ref{fig:darts-l2-ops-alpha} give us such an illustration, where $L_{0-1}$ has the same weight $w=10$ as in Fair DARTS. Not surprisingly, under the exclusive competition by softmax, skip connections exploit even more from unfair advantages as we drive the weak-performing operations towards zero (Fig.~\ref{fig:darts-l2-ops-alpha}). Noticeably,  there is a \emph{domino effect}, as the weakest operation decrease its weight, the second weakest follows, so on and so forth. This effect speeds up the aggregation and as a result, more skip connections stand out (Fig.~\ref{fig:darts-l2-skip-no-acc}). As the rest better-performing operations are contending with each other, it still finds difficulty to determine which one is the best. 
Besides, training its inferred best models reaches 96.77 $\pm$ 0.29\% on CIFAR-10  (run 7 times each with different seeds), which is not too different from the original DARTS. Therefore, we conclude that  applying $L_{0-1}$ alone is not enough to solve the existing problems in DARTS. In fact, $L_{0-1}$ cannot handle the softmax case inherently.

\begin{table*}
	\begin{center}
		\caption{Fair DARTS architecture genotypes.  See FairDARTS-a in Fig.~\ref{fig:normal-reduce-architecture}}
		\label{tab:fairdarts_models}
		\begin{footnotesize}
			\begin{tabular}{cp{10cm}}
				\hline
				Model & Architecture Genotype  \\
				\hline
				FairDARTS-b & 
				Genotype(normal=[('sep\_conv\_3x3', 2, 0), ('sep\_conv\_3x3', 2, 1), ('sep\_conv\_3x3', 3, 1), ('dil\_conv\_3x3', 4, 0), ('sep\_conv\_5x5', 4, 1), ('dil\_conv\_5x5', 5, 1)], normal\_concat=range(2, 6), 
				reduce=[('skip\_connect', 2, 0), ('dil\_conv\_3x3', 2, 1), ('skip\_connect', 3, 0), ('dil\_conv\_3x3', 3, 1), ('max\_pool\_3x3', 4, 0), ('sep\_conv\_3x3', 4, 1), ('skip\_connect', 5, 2), ('max\_pool\_3x3', 5, 0)], reduce\_concat=range(2, 6))

				\\
				
				FairDARTS-c & 	Genotype(normal=[('max\_pool\_3x3', 2, 0), ('sep\_conv\_5x5', 2, 1), ('dil\_conv\_3x3', 4, 0), ('dil\_conv\_5x5', 4, 2), ('skip\_connect', 5, 3), ('sep\_conv\_3x3', 5, 0)], normal\_concat=range(2, 6),
				reduce=[('dil\_conv\_3x3', 2, 1), ('dil\_conv\_5x5', 2, 0), ('dil\_conv\_3x3', 3, 0), ('sep\_conv\_3x3', 3, 1), ('sep\_conv\_5x5', 4, 0), ('sep\_conv\_5x5', 4, 3), ('sep\_conv\_5x5', 5, 0), ('skip\_connect', 5, 1)], reduce\_concat=range(2, 6))
				
				\\
				FairDARTS-d &  				
				Genotype(normal=[('sep\_conv\_3x3', 2, 0), ('sep\_conv\_5x5', 2, 1), ('dil\_conv\_3x3', 3, 1), ('max\_pool\_3x3', 3, 0), ('dil\_conv\_3x3', 4, 0), ('dil\_conv\_3x3', 4, 1), ('sep\_conv\_3x3', 5, 0), ('dil\_conv\_5x5', 5, 1)], normal\_concat=range(2, 6), 
				reduce=[('max\_pool\_3x3', 2, 0), ('sep\_conv\_5x5', 2, 1), ('avg\_pool\_3x3', 3, 0), ('dil\_conv\_5x5', 3, 2), ('dil\_conv\_3x3', 4, 3), ('avg\_pool\_3x3', 4, 0), ('avg\_pool\_3x3', 5, 0), ('skip\_connect', 5, 3)], reduce\_concat=range(2, 6))
				
				\\
				FairDARTS-e & Genotype(normal=[('sep\_conv\_3x3', 2, 0), ('sep\_conv\_3x3', 2, 1), ('dil\_conv\_3x3', 4, 1), ('dil\_conv\_3x3', 4, 2), ('dil\_conv\_3x3', 5, 0), ('dil\_conv\_5x5', 5, 1)], normal\_concat=range(2, 6), 
				reduce=[('max\_pool\_3x3', 2, 1), ('max\_pool\_3x3', 2, 0), ('max\_pool\_3x3', 3, 1), ('max\_pool\_3x3', 3, 0), ('sep\_conv\_5x5', 4, 1), ('max\_pool\_3x3', 4, 0), ('avg\_pool\_3x3', 5, 0), ('dil\_conv\_5x5', 5, 1)], reduce\_concat=range(2, 6))
				
				\\
				FairDARTS-f &  Genotype(normal=[('max\_pool\_3x3', 2, 0), ('sep\_conv\_3x3', 2, 1), ('dil\_conv\_3x3', 3, 1), ('sep\_conv\_5x5', 4, 1), ('sep\_conv\_3x3', 5, 0), ('sep\_conv\_3x3', 5, 1)], normal\_concat=range(2, 6), 
				reduce=[('max\_pool\_3x3', 2, 0), ('max\_pool\_3x3', 2, 1), ('max\_pool\_3x3', 3, 0), ('dil\_conv\_3x3', 3, 1), ('dil\_conv\_3x3', 4, 2), ('max\_pool\_3x3', 4, 0), ('max\_pool\_3x3', 5, 0), ('sep\_conv\_3x3', 5, 1)], reduce\_concat=range(2, 6))
				
				\\
				FairDARTS-g & Genotype(normal=[('sep\_conv\_3x3', 2, 0), ('sep\_conv\_3x3', 2, 1), ('skip\_connect', 4, 3), ('sep\_conv\_5x5', 4, 1), ('dil\_conv\_3x3', 5, 0), ('sep\_conv\_3x3', 5, 1)], normal\_concat=range(2, 6), 
				reduce=[('avg\_pool\_3x3', 2, 1), ('skip\_connect', 2, 0), ('skip\_connect', 3, 2), ('max\_pool\_3x3', 3, 1), ('sep\_conv\_5x5', 4, 3), ('max\_pool\_3x3', 4, 0), ('dil\_conv\_3x3', 5, 1), ('dil\_conv\_3x3', 5, 4)], reduce\_concat=range(2, 6))
				\\ 
				\hline
			\end{tabular}
		\end{footnotesize}
	\end{center}
\end{table*}

\begin{table*}
	\begin{center}
		\caption{Architecture genotypes when adding Gaussian noise to DARTS. Discussed in Section \ref{sec:discussion} (main text)}
		\label{tab:noisy-darts-lr-decay}
		\begin{footnotesize}
			\begin{tabular}{cp{10cm}}  
				\hline
				Model & Architecture Genotype  \\
				\hline
				0 &  Genotype(normal=[('sep\_conv\_3x3', 1), ('sep\_conv\_3x3', 0), ('dil\_conv\_5x5', 1), ('dil\_conv\_3x3', 2), ('sep\_conv\_3x3', 1), ('dil\_conv\_5x5', 3), ('dil\_conv\_5x5', 4), ('dil\_conv\_3x3', 3)], normal\_concat=range(2, 6),
				reduce=[('max\_pool\_3x3', 0), ('max\_pool\_3x3', 1), ('avg\_pool\_3x3', 0), ('dil\_conv\_5x5', 2), ('skip\_connect', 1), ('max\_pool\_3x3', 0), ('skip\_connect', 3), ('avg\_pool\_3x3', 1)], reduce\_concat=range(2, 6))
				
				\\
				1 & Genotype(normal=[('dil\_conv\_3x3', 1), ('sep\_conv\_3x3', 0), ('dil\_conv\_3x3', 2), ('dil\_conv\_3x3', 1), ('sep\_conv\_3x3', 1), ('dil\_conv\_5x5', 3), ('dil\_conv\_3x3', 3), ('dil\_conv\_5x5', 4)], normal\_concat=range(2, 6),
				reduce=[('dil\_conv\_5x5', 1), ('skip\_connect', 0), ('max\_pool\_3x3', 0), ('dil\_conv\_3x3', 2), ('skip\_connect', 3), ('skip\_connect', 2), ('skip\_connect', 2), ('skip\_connect', 3)], reduce\_concat=range(2, 6))
				
				\\
				2 & Genotype(normal=[('dil\_conv\_3x3', 1), ('sep\_conv\_3x3', 0), ('dil\_conv\_3x3', 1), ('dil\_conv\_3x3', 2), ('sep\_conv\_3x3', 1), ('dil\_conv\_3x3', 3), ('dil\_conv\_3x3', 4), ('dil\_conv\_3x3', 1)], normal\_concat=range(2, 6),
				reduce=[('max\_pool\_3x3', 0), ('dil\_conv\_3x3', 1), ('avg\_pool\_3x3', 0), ('dil\_conv\_5x5', 2), ('max\_pool\_3x3', 0), ('skip\_connect', 2), ('skip\_connect', 2), ('avg\_pool\_3x3', 0)], reduce\_concat=range(2, 6))
				
				\\
				3 & Genotype(normal=[('dil\_conv\_3x3', 1), ('skip\_connect', 0), ('dil\_conv\_3x3', 2), ('dil\_conv\_3x3', 1), ('sep\_conv\_3x3', 1), ('dil\_conv\_3x3', 3), ('dil\_conv\_3x3', 4), ('dil\_conv\_5x5', 2)], normal\_concat=range(2, 6),
				reduce=[('avg\_pool\_3x3', 0), ('dil\_conv\_5x5', 1), ('avg\_pool\_3x3', 0), ('dil\_conv\_5x5', 2), ('avg\_pool\_3x3', 0), ('dil\_conv\_5x5', 3), ('dil\_conv\_5x5', 4), ('avg\_pool\_3x3', 0)], reduce\_concat=range(2, 6))
				\\ 
				\hline
			\end{tabular}
		\end{footnotesize}
	\end{center}
	
\end{table*}

\begin{table*}
	\begin{center}
		\caption{Randomly sampled architecture genotypes in $S_1$ setting $M=2$. Discussed in Section \ref{sec:discussion} (main text)}
		\label{tab:random-m2noflops}
		\begin{footnotesize}
			\begin{tabular}{cp{11cm}}  
				\hline
				Model & Architecture Genotype  \\
				\hline
				0 & Genotype(normal=[('skip\_connect', 1), ('sep\_conv\_5x5', 0), ('skip\_connect', 1), ('dil\_conv\_5x5', 0), ('avg\_pool\_3x3', 1), ('dil\_conv\_3x3', 0), ('max\_pool\_3x3', 0), ('sep\_conv\_3x3', 3)], normal\_concat=range(2, 6),
				reduce=[('dil\_conv\_3x3', 1), ('dil\_conv\_5x5', 0), ('max\_pool\_3x3', 1), ('dil\_conv\_5x5', 2), ('skip\_connect', 0), ('dil\_conv\_3x3', 1), ('avg\_pool\_3x3', 4), ('sep\_conv\_5x5', 1)], reduce\_concat=range(2, 6))
				
				\\
				1 & Genotype(normal=[('skip\_connect', 1), ('skip\_connect', 0), ('dil\_conv\_3x3', 0), ('sep\_conv\_5x5', 2), ('dil\_conv\_5x5', 1), ('sep\_conv\_5x5', 3), ('dil\_conv\_3x3', 3), ('max\_pool\_3x3', 4)], normal\_concat=range(2, 6),
				reduce=[('sep\_conv\_3x3', 1), ('sep\_conv\_3x3', 0), ('dil\_conv\_3x3', 2), ('max\_pool\_3x3', 1), ('sep\_conv\_3x3', 0), ('sep\_conv\_3x3', 1), ('max\_pool\_3x3', 2), ('skip\_connect', 3)], reduce\_concat=range(2, 6))
				
				\\
				2 &  Genotype(normal=[('avg\_pool\_3x3', 0), ('skip\_connect', 1), ('dil\_conv\_5x5', 0), ('dil\_conv\_3x3', 2), ('sep\_conv\_5x5', 2), ('skip\_connect', 3), ('avg\_pool\_3x3', 1), ('avg\_pool\_3x3', 4)], normal\_concat=range(2, 6),
				reduce=[('avg\_pool\_3x3', 0), ('avg\_pool\_3x3', 1), ('avg\_pool\_3x3', 0), ('sep\_conv\_5x5', 1), ('sep\_conv\_3x3', 2), ('avg\_pool\_3x3', 1), ('sep\_conv\_3x3', 2), ('max\_pool\_3x3', 1)], reduce\_concat=range(2, 6))
				
				\\
				3& Genotype(normal=[('skip\_connect', 1), ('dil\_conv\_3x3', 0), ('sep\_conv\_3x3', 0), ('avg\_pool\_3x3', 2), ('dil\_conv\_5x5', 0), ('dil\_conv\_3x3', 3), ('max\_pool\_3x3', 1), ('skip\_connect', 4)], normal\_concat=range(2, 6),
				reduce=[('dil\_conv\_3x3', 0), ('sep\_conv\_5x5', 1), ('sep\_conv\_5x5', 1), ('sep\_conv\_5x5', 0), ('avg\_pool\_3x3', 0), ('sep\_conv\_5x5', 2), ('sep\_conv\_3x3', 0), ('max\_pool\_3x3', 1)], reduce\_concat=range(2, 6))
				
				\\
				4 & Genotype(normal=[('dil\_conv\_5x5', 1), ('skip\_connect', 0), ('sep\_conv\_5x5', 0), ('skip\_connect', 2), ('sep\_conv\_5x5', 3), ('dil\_conv\_5x5', 2), ('avg\_pool\_3x3', 0), ('max\_pool\_3x3', 3)], normal\_concat=range(2, 6),
				reduce=[('dil\_conv\_3x3', 0), ('dil\_conv\_3x3', 1), ('max\_pool\_3x3', 0), ('avg\_pool\_3x3', 2), ('max\_pool\_3x3', 0), ('avg\_pool\_3x3', 2), ('dil\_conv\_3x3', 0), ('dil\_conv\_3x3', 2)], reduce\_concat=range(2, 6))
				
				\\
				5 & Genotype(normal=[('sep\_conv\_5x5', 0), ('skip\_connect', 1), ('sep\_conv\_3x3', 2), ('sep\_conv\_3x3', 0), ('avg\_pool\_3x3', 2), ('skip\_connect', 0), ('sep\_conv\_3x3', 0), ('dil\_conv\_5x5', 3)], normal\_concat=range(2, 6),
				reduce=[('avg\_pool\_3x3', 1), ('sep\_conv\_5x5', 0), ('dil\_conv\_3x3', 2), ('skip\_connect', 1), ('avg\_pool\_3x3', 2), ('skip\_connect', 1), ('sep\_conv\_3x3', 4), ('dil\_conv\_3x3', 1)], reduce\_concat=range(2, 6))
				
				\\
				6 &  Genotype(normal=[('skip\_connect', 1), ('sep\_conv\_5x5', 0), ('max\_pool\_3x3', 0), ('dil\_conv\_3x3', 1), ('sep\_conv\_5x5', 1), ('avg\_pool\_3x3', 0), ('skip\_connect', 4), ('sep\_conv\_5x5', 0)], normal\_concat=range(2, 6),
				reduce=[('dil\_conv\_5x5', 1), ('sep\_conv\_3x3', 0), ('sep\_conv\_5x5', 0), ('dil\_conv\_5x5', 1), ('max\_pool\_3x3', 1), ('max\_pool\_3x3', 3), ('dil\_conv\_5x5', 2), ('max\_pool\_3x3', 1)], reduce\_concat=range(2, 6))
				\\
				\hline
			\end{tabular}
		\end{footnotesize}
	\end{center}
\end{table*}

\begin{table*}
	\begin{center}
		\caption{Randomly sampled architecture genotypes in $S_1$ setting $M=2$ and multiply-adds $>$ 500M. Discussed in Section \ref{sec:discussion} (main text)}
		\label{tab:random-m2500flops}
		\begin{footnotesize}
			\begin{tabular}{cp{11cm}}  
				\hline
				Model & Architecture Genotype  \\
				\hline
				0 & Genotype(normal=[('skip\_connect', 1), ('sep\_conv\_3x3', 0), ('skip\_connect', 0), ('dil\_conv\_3x3', 2), ('sep\_conv\_3x3', 1), ('dil\_conv\_5x5', 2), ('sep\_conv\_3x3', 2), ('dil\_conv\_5x5', 1)], normal\_concat=range(2, 6),
				reduce=[('skip\_connect', 0), ('sep\_conv\_5x5', 1), ('avg\_pool\_3x3', 2), ('dil\_conv\_5x5', 0), ('sep\_conv\_3x3', 0), ('dil\_conv\_5x5', 2), ('sep\_conv\_5x5', 1), ('dil\_conv\_5x5', 2)], reduce\_concat=range(2, 6))
				
				\\
				1 &  Genotype(normal=[('dil\_conv\_3x3', 0), ('sep\_conv\_3x3', 1), ('skip\_connect', 2), ('skip\_connect', 1), ('sep\_conv\_5x5', 0), ('dil\_conv\_3x3', 1), ('sep\_conv\_5x5', 1), ('dil\_conv\_5x5', 4)], normal\_concat=range(2, 6),
				reduce=[('dil\_conv\_3x3', 0), ('dil\_conv\_3x3', 1), ('avg\_pool\_3x3', 1), ('skip\_connect', 0), ('dil\_conv\_5x5', 1), ('skip\_connect', 3), ('skip\_connect', 0), ('max\_pool\_3x3', 3)], reduce\_concat=range(2, 6))
				
				\\
				2 & Genotype(normal=[('sep\_conv\_5x5', 0), ('max\_pool\_3x3', 1), ('sep\_conv\_5x5', 1), ('skip\_connect', 0), ('skip\_connect', 2), ('max\_pool\_3x3', 1), ('sep\_conv\_5x5', 3), ('sep\_conv\_3x3', 2)], normal\_concat=range(2, 6),
				reduce=[('sep\_conv\_5x5', 0), ('skip\_connect', 1), ('max\_pool\_3x3', 1), ('sep\_conv\_5x5', 2), ('dil\_conv\_5x5', 3), ('max\_pool\_3x3', 0), ('dil\_conv\_3x3', 1), ('max\_pool\_3x3', 2)], reduce\_concat=range(2, 6))
				
				\\
				3 & Genotype(normal=[('sep\_conv\_3x3', 0), ('max\_pool\_3x3', 1), ('dil\_conv\_5x5', 2), ('dil\_conv\_5x5', 0), ('sep\_conv\_5x5', 3), ('sep\_conv\_5x5', 0), ('skip\_connect', 3), ('skip\_connect', 0)], normal\_concat=range(2, 6),
				reduce=[('avg\_pool\_3x3', 0), ('sep\_conv\_5x5', 1), ('avg\_pool\_3x3', 1), ('dil\_conv\_3x3', 0), ('dil\_conv\_5x5', 3), ('sep\_conv\_5x5', 2), ('avg\_pool\_3x3', 1), ('dil\_conv\_5x5', 4)], reduce\_concat=range(2, 6))
				
				\\
				4 &  Genotype(normal=[('dil\_conv\_5x5', 0), ('sep\_conv\_5x5', 1), ('sep\_conv\_3x3', 2), ('skip\_connect', 0), ('sep\_conv\_5x5', 3), ('sep\_conv\_5x5', 0), ('avg\_pool\_3x3', 1), ('skip\_connect', 0)], normal\_concat=range(2, 6),
				reduce=[('max\_pool\_3x3', 1), ('skip\_connect', 0), ('dil\_conv\_5x5', 1), ('dil\_conv\_5x5', 2), ('sep\_conv\_5x5', 0), ('sep\_conv\_3x3', 1), ('avg\_pool\_3x3', 1), ('skip\_connect', 0)], reduce\_concat=range(2, 6))
				
				\\
				5 & Genotype(normal=[('sep\_conv\_5x5', 1), ('sep\_conv\_5x5', 0), ('skip\_connect', 2), ('sep\_conv\_3x3', 0), ('dil\_conv\_5x5', 2), ('dil\_conv\_3x3', 3), ('max\_pool\_3x3', 0), ('skip\_connect', 2)], normal\_concat=range(2, 6),
				reduce=[('max\_pool\_3x3', 0), ('dil\_conv\_5x5', 1), ('max\_pool\_3x3', 2), ('skip\_connect', 0), ('dil\_conv\_5x5', 0), ('sep\_conv\_5x5', 3), ('sep\_conv\_3x3', 1), ('dil\_conv\_5x5', 3)], reduce\_concat=range(2, 6))
				
				\\
				6 & Genotype(normal=[('avg\_pool\_3x3', 0), ('sep\_conv\_5x5', 1), ('sep\_conv\_5x5', 0), ('skip\_connect', 1), ('dil\_conv\_5x5', 0), ('sep\_conv\_5x5', 1), ('skip\_connect', 4), ('sep\_conv\_5x5', 0)], normal\_concat=range(2, 6),
				reduce=[('dil\_conv\_5x5', 1), ('avg\_pool\_3x3', 0), ('avg\_pool\_3x3', 2), ('sep\_conv\_5x5', 0), ('max\_pool\_3x3', 0), ('dil\_conv\_5x5', 2), ('sep\_conv\_5x5', 3), ('skip\_connect', 0)], reduce\_concat=range(2, 6))
				\\
				\hline
			\end{tabular}
		\end{footnotesize}
	\end{center}
\end{table*}
\clearpage
\bibliographystyle{splncs04}
\bibliography{egbib}
\end{document}